\documentclass[10pt,twocolumn,letterpaper]{article}
\usepackage{color}
\usepackage[table,dvipsnames]{xcolor}
\usepackage{bm}
\usepackage{times}
\usepackage{epsfig}
\usepackage{graphicx}
\usepackage{amsmath}
\usepackage{amssymb}
\newcounter{RNum}
\usepackage{booktabs}
\renewcommand{\theRNum}{\arabic{RNum}}
\newcommand{\Remark}{\noindent\textit{\textbf{Remark}~\refstepcounter{RNum}\textbf{\theRNum}: }}
\usepackage{multirow}
\usepackage{float}
\usepackage{graphics} 
\usepackage{epsfig} 
\usepackage{mathrsfs}
\usepackage{times} 
\usepackage{amsmath} 
\usepackage{amssymb}  

\newcommand{\NoOne}[1]{\textcolor{red}{#1}}

\usepackage{cvpr}              

\usepackage{graphicx}
\usepackage{amsmath}
\usepackage{amssymb}
\usepackage{booktabs}
\newcommand{\Trackname}{TCTrack}

\usepackage{amsfonts}

\usepackage{graphicx}
\usepackage{pifont}

\usepackage[pagebackref,breaklinks,colorlinks]{hyperref}

\usepackage[capitalize]{cleveref}
\crefname{section}{Sec.}{Secs.}
\Crefname{section}{Section}{Sections}
\Crefname{table}{Table}{Tables}
\crefname{table}{Tab.}{Tabs.}

\def\cvprPaperID{2734} 
\def\confName{CVPR}
\def\confYear{2022}

\begin{document}
	
	\title{TCTrack: Temporal Contexts for Aerial Tracking}
	
	\author{Ziang Cao$^{1}$, Ziyuan Huang$^{2}$, Liang Pan$^{3}$, Shiwei Zhang$^{4}$, Ziwei Liu$^{3}$, Changhong Fu$^{1,}\thanks{Corresponding author}$~\\
		$^{1}$Tongji University $^{2}$National University of Singapore $^{3}$S-Lab, Nanyang Technological University \\$^{4}$DAMO Academy, Alibaba Group\\
		{\tt\small caoang233@gmail.com, ziyuan.huang@u.nus.edu, \{liang.pan, ziwei.liu\}@ntu.edu.sg}\\{\tt\small zhangjin.zsw@alibaba-inc.com changhongfu@tongji.edu.cn}
	}
	
	\maketitle
	
	\begin{abstract}

		Temporal contexts among consecutive frames are far from being fully utilized in existing visual trackers. In this work, we present \textbf{TCTrack}\footnote{\url{https://github.com/vision4robotics/TCTrack}}, a comprehensive framework to fully exploit temporal contexts for aerial tracking. The temporal contexts are incorporated at \textbf{two levels}: the extraction of \textbf{features} and the refinement of \textbf{similarity maps}. Specifically, for feature extraction, an online temporally adaptive convolution is proposed to enhance the spatial features using temporal information, which is achieved by dynamically calibrating the convolution weights according to the previous frames. For similarity map refinement, we propose an adaptive temporal transformer, which first effectively encodes temporal knowledge in a memory-efficient way, before the temporal knowledge is decoded for accurate adjustment of the similarity map. TCTrack is effective and efficient: evaluation on four aerial tracking benchmarks shows its impressive performance; real-world UAV tests show its high speed of over 27 FPS on NVIDIA Jetson AGX Xavier.

	\end{abstract}

	\section{Introduction}
	\label{sec:intro}
	
	Visual tracking is one of the most fundamental tasks in computer vision. Owing to the superior mobility of unmanned aerial vehicles (UAVs), tracking-based applications are experiencing rapid developments, \textit{e.g.}, motion object analysis~\cite{8736008}, geographical survey~\cite{5942155}, and visual localization~\cite{Li_2020_CVPR}. 
	Nevertheless, aerial tracking still faces two difficulties: 1) aerial conditions inevitably introduce special challenges including motion blur, camera motion, occlusion, \textit{etc}; 2) the limited power of aerial platforms restricts the computational resource, impeding the deployment of time-consuming state-of-the-art methods~\cite{cao2021iccv}. Hence, an ideal tracker for aerial tracking must be robust and efficient.
	
	Most existing trackers adopt the standard tracking-by-detection framework and perform detection for each frame independently. Among these trackers, discriminative correlation filter (DCF)-based methods are widely applied on aerial platforms because of their high efficiency and low resource requirements originated from the operations in the Fourier domain~\cite{kiani2017learning,kcf,SRDCF}.  
	However, these trackers struggle when there are fast motions and severe appearance variations.
	Recently, the Siamese-based network has emerged as a strong framework for accurate and robust tracking~\cite{siamfc,siamrpn,8953466,9010649,8954116}.
	Its efficiency is also optimized in \cite{fu2020siamese,cao2021siamapn,9477413} for the real-time deployment of Siamese-based trackers on aerial platforms.
	
	
	\begin{figure}[t]
		\centering

		\includegraphics[width=0.47\textwidth]{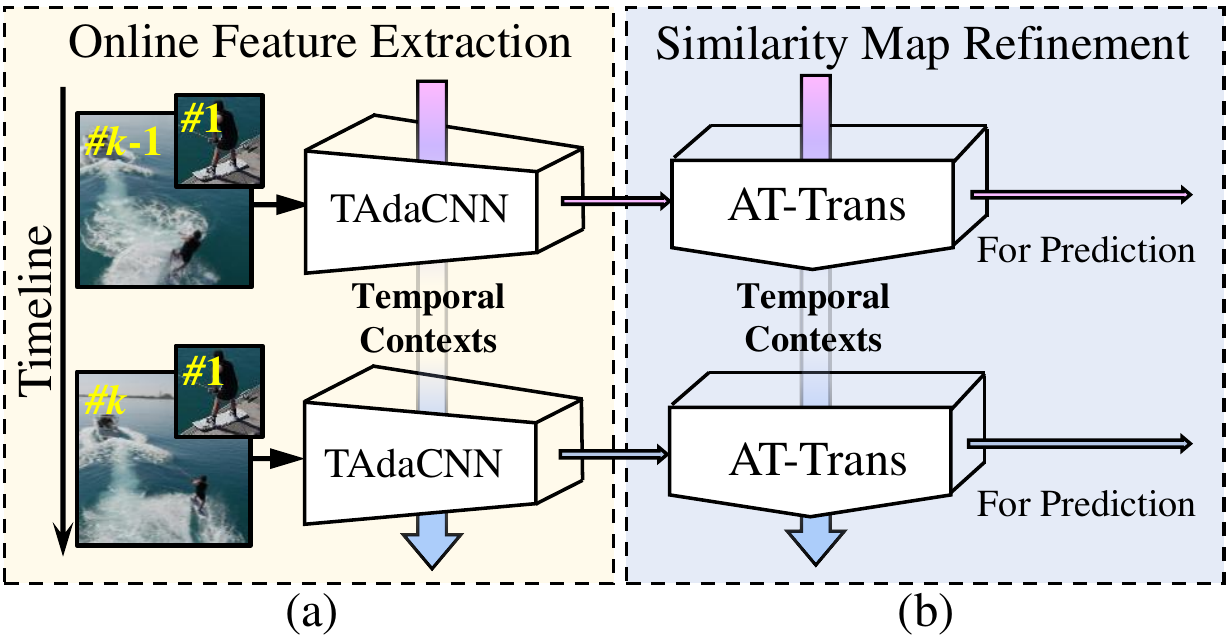}
		\vspace{-6pt}
		\caption{
			Overview of our framework namely TCTrack. It exploits temporal information at two levels: (a) \textit{the extraction of features} by the temporally adaptive convolutional neural networks (TAdaCNN) and (b) \textit{the refinement of similarity maps} by the adaptive temporal transformer (AT-Trans). 
		}
		\vspace{-18pt}
		\label{fig:page1}
		
	\end{figure}
	
	However, the strong correlations inherently existing among consecutive frames, \textit{i.e.,} the temporal information, are neglected by these frameworks, which makes it difficult for these approaches to perceive the motion information of the target objects.
	Therefore, those trackers are more likely to fail when the target undergoes severe appearance change caused by different complex conditions such as large motions and occlusions.
	This has sparked the recent research into how to make use of temporal information for visual tracking.
	For DCF-based approaches, the variation in the response maps along the temporal dimension is penalized~\cite{Huang2019ICCV,Li_2020_CVPR}, which guides the current response map by previous ones.
	In Siamese-based networks, which is the focus of this work, temporal information is introduced in most works through dynamic templates, which integrates historical object appearance in the current template through concatenation~\cite{yan2021learning}, weighted sum~\cite{updatenet}, graph network~\cite{gct}, transformer~\cite{wang2021transformer}, or memory networks~\cite{fu2021stmtrack,yang2018learning}. 
	Despite their success in introducing temporal information into the visual tracking task, most of the explorations are restricted to \textit{only a single stage}, \textit{i.e.,} the template feature, in the whole tracking pipeline.
	

	
	In this work, we present a comprehensive framework for exploiting temporal contexts in Siamese-based networks, which we call \textbf{TCTrack}. As shown in Fig.~\ref{fig:page1}, TCTrack introduces temporal context into the tracking pipeline \textit{at two levels}, \textit{i.e., }features and similarity maps. 
	At the \textbf{feature level}, we propose an online temporally adaptive convolution (TAdaConv), where features are extracted with convolution weights dynamically calibrated by the previous frames. 
	Based on this operation, we transform the standard convolutional networks to temporally adaptive ones (TAdaCNN). 
	Since the calibration in the online TAdaConv is based on the global descriptor of the features in the previous frames, TAdaCNN only introduces a negligible frame rate drop but notably improves the tracking performance. 
	At the \textbf{similarity map level}, an adaptive temporal transformer (AT-Trans) is proposed to refine the similarity map according to the temporal information. Specifically, AT-Trans adopts an encoder-decoder structure, where \textbf{(i)} the encoder produces the temporal prior knowledge for the current time step, by integrating the previous prior with the current similarity map, and \textbf{(ii)} the decoder refines the similarity map based on the produced temporal prior knowledge in an adaptive way. 
	Compared to \cite{fu2021stmtrack,wang2021transformer,gct}, AT-Trans is memory efficient and thus edge-platform friendly since we keep updating the temporal prior knowledge at each frame. 
	Overall, our approach provides a holistic temporal encoding framework to handle temporal contexts in Siamese-based aerial tracking. 
	

	Extensive evaluations of TCTrack show both the effectiveness and the efficiency of the proposed framework.
	Competitive accuracy and precision are observed on four standard aerial tracking benchmarks in comparison with 51 state-of-the-art trackers, where TCTrack also has a high frame rate of 125.6 FPS on PC. Real-world deployment on NVIDIA Jetson AGX Xavier shows that TCTrack maintains impressive stability and robustness for aerial tracking, running at a frame rate of over 27 FPS.

	\section{Related Work}\label{sec:Related Work}
	
	\begin{figure*}[t]
		\centering

		\includegraphics[width=1\textwidth]{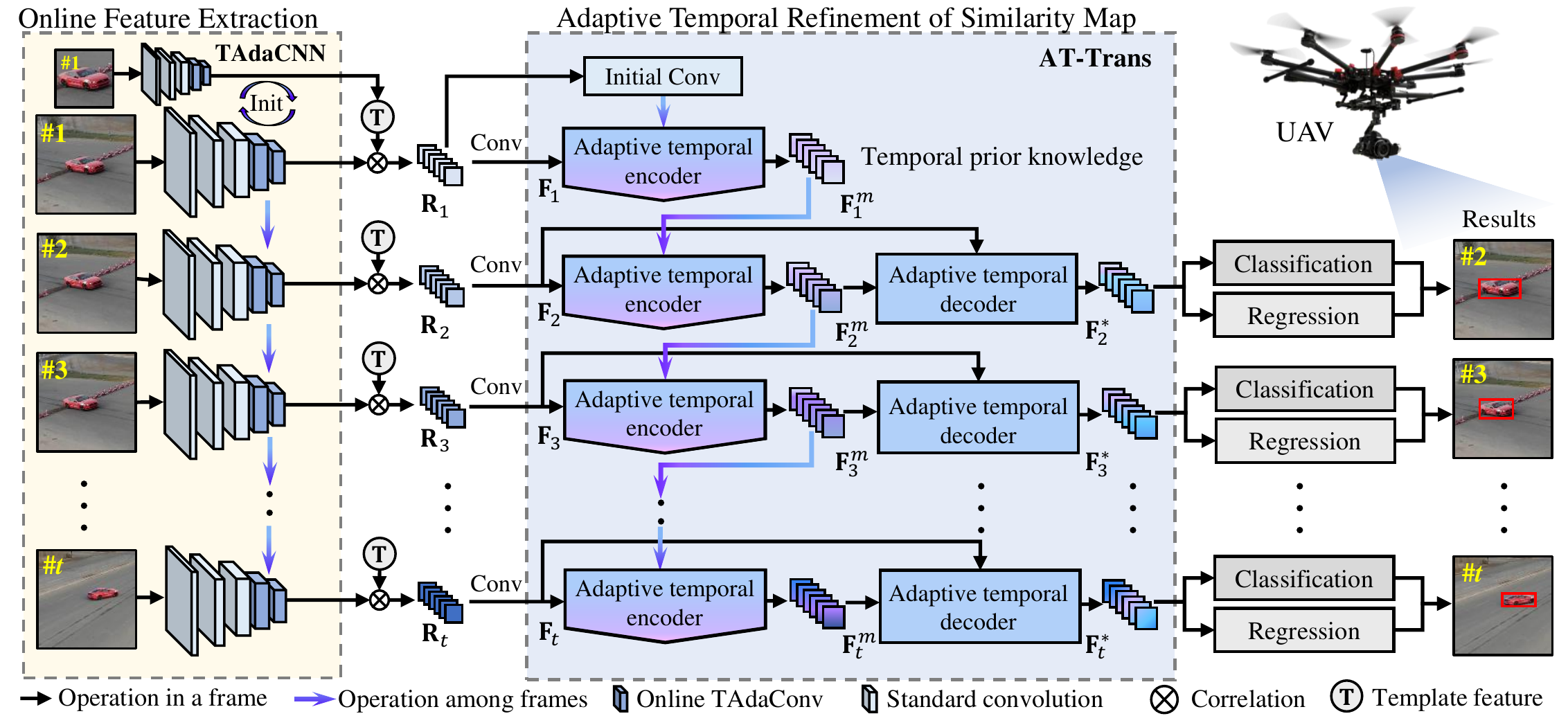}
		\vspace{-20pt}
		\caption{Overview of our framework. It mainly consists of three components, \textit{i.e.}, TAdaCNN for online feature extraction shown in Fig.~\ref{fig:conv}, AT-Trans for similarity map refinement shown in Fig.~\ref{fig:work}, and classification\&regression for final prediction. This figure illustrates the workflow of our TCTrack when tracking sequences are \textit{t} frames. Through temporal contexts before correlation and after, comprehensive temporal knowledge is introduced in our framework. Best view in color.}
		\vspace{-10pt}
		\label{fig:1}
	\end{figure*}

	\noindent\textbf{Tracking by detection.} After D. S. Bolme \textit{et al.} firstly proposed the MOSSE filter~\cite{Bolme2010CVPR}, many researches~\cite{kcf,SRDCF,kiani2017learning} have been made to boost the tracking performance. However, since they suffer from poor representative feature expression, they are hard to maintain robustness under complex aerial tracking conditions. Recently, Siamese-based trackers have stood out attributing to their SOTA accuracy and attractive efficiency~\cite{siamfc,siamrpn,siamdw,8954116,9157720,chen2020siamese,bhat2020know}. For meeting the aerial tracking requirement, some works propose efficient tracking methods~\cite{fu2020siamese,cao2021siamapn,9477413}. 
	
	Despite achieving SOTA performance, those trackers above disregard the temporal contexts in the tracking scenarios, thereby blocking the performance improvement. Differently, our tracker can effectively model the historical temporal contexts during the tracking for increasing the discriminability and robustness.

	\noindent\textbf{Temporal-based tracking methods.}
	Previously, many works are devoted to exploiting the temporal information in tracking scenarios for raising the tracking performance~\cite{Huang2019ICCV,Li_2020_CVPR,dai2019visual,8578613}. Recently, many DL-based temporal tracking methods focus on dynamic templates based on transformer integration~\cite{wang2021transformer}, template memory update~\cite{yang2018learning,fu2021stmtrack,dsiam}, graph network~\cite{gct}, weighted sum~\cite{updatenet}, and explicit template update~\cite{yan2021learning}. They try to update the template features in an explicit way or implicit way based on the pre-defined parameters. Then, based on the transformed template features, those trackers exploit the discrete temporal information in tracking sequences.
	
	Despite superior tracking performance, they introduce temporal information via only a single level in the whole tracking pipeline, blocking further improvement of tracking performance. To fully exploit the temporal contexts, in this work, we propose a comprehensive framework for exploring the temporal contexts via two levels, \textit{i.e.}, features level and similarity maps level.



	\noindent\textbf{Temporal modelling in videos.}
	Modelling the temporal dynamics is essential for a genuine understanding of videos. Hence, it is widely explored in both supervised~\cite{feichtenhofer2019slowfast,tran2018r2p1d,huang2021tadaconv,liu2021tam,wang2018nonlocal,lin2019tsm} and self-supervised paradigm~\cite{huang2021mosi,han2019dpc,han2020memdpc,kim2019stpuzzle,huo2021csj}.
	Self-supervised approaches learns temporal modelling by solving various pre-text tasks, such as dense future prediction~\cite{han2019dpc,han2020memdpc}, jigsaw puzzle solving~\cite{kim2019stpuzzle,huo2021csj}, and pseudo motion classification~\cite{huang2021mosi}, \textit{etc.}
	Supervised video recognition explores various connections between different frames, such as 3D convolutions~\cite{tran2015c3d}, temporal convolution~\cite{tran2018r2p1d}, and temporal shift~\cite{lin2019tsm}, \textit{etc.}
	Closely related to our work is the temporally adaptive convolutions~\cite{huang2021tadaconv}, which is applied for temporal modeling in videos. In this work, to adapt to the tracking task, we propose an online CNN which can extract spatial features according to temporal contexts for enriching the temporal information comprehensively.

	
	\section{Temporal Contexts for Aerial Tracking}\label{sec:Proposed Method}
	In this section, the detailed structure of our framework is described as shown in Fig.~\ref{fig:1}.
	The proposed framework considers temporal contexts from two new perspectives: \textbf{(1)} online feature extraction where we incorporate temporal context by TAdaCNN (Sec.~\ref{feat}); and \textbf{(2)} similarity map refinement where we use a novel AT-Trans to encode the temporal knowledge and then refine the similarity map according to the temporal prior knowledge (Sec.~\ref{tf}).

	\subsection{Feature extraction with online TAdaConv}
	\label{feat}

\begin{figure}[t]
		\centering

		\includegraphics[width=0.48\textwidth]{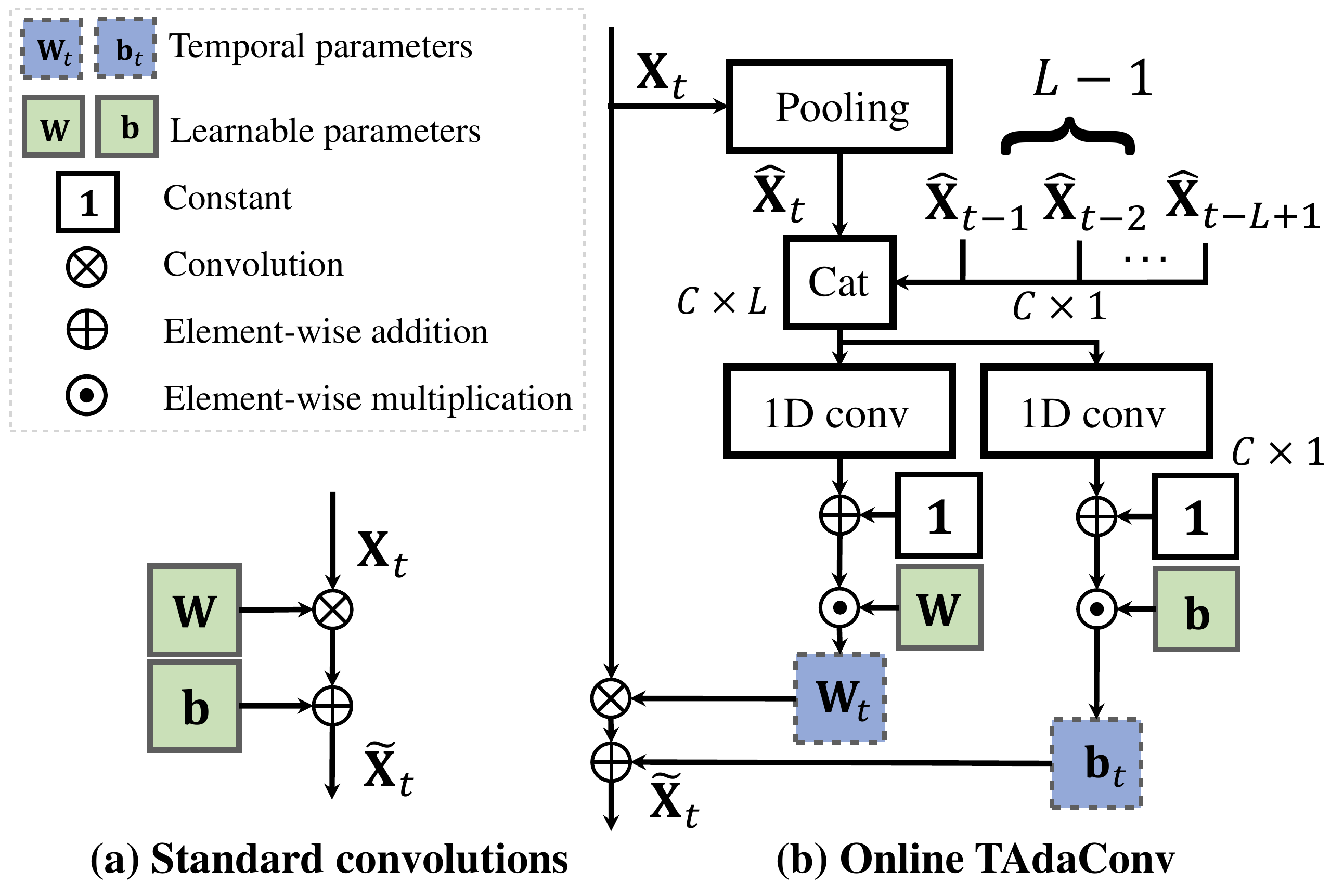}
		\vspace{-20pt}
		\caption{The schema of our online TAdaConv. The temporal calibration factor is generated by the feature sequences (number of its is L). Based on the temporal vectors, the parameters of our online TAdaConv can be adjusted adaptively in every frame.}
		\label{fig:conv}
		\vspace{-10pt}
	\end{figure}

	As a key component of our framework, an online TAdaConv is proposed for feature extraction based on~\cite{huang2021tadaconv} to consider temporal contexts whose structures are shown in Fig.~\ref{fig:conv}. Formally, given the input feature to the online TAdaConv at a certain stage in the network $\mathbf{X}_t$ in the $t$-th frame, the output of the online TAdaConv $\mathbf{\tilde{X}}_t$ can be obtained as follows: 
	\begin{equation}
		\mathbf{\tilde{X}}_t = \mathbf{W}_t * \mathbf{X}_t + \mathbf{b}_t\ ,
	\end{equation}
	\noindent where the operator $*$ denotes the convolution operation and $\mathbf{W}_t, \mathbf{b}_t$ are the temporal weight and bias of our convolution. A standard convolution layer uses learnable parameters for weights and bias, and shares them in the whole tracking sequence. Differently, in our online convolution layer, the parameters are calculated by the learnable parameters ($\mathbf{W}_b$ and $\mathbf{b}_b$) and calibration factors that are varied for each frame, \textit{i.e.,} $\mathbf{W}_t=\mathbf{W}_b\cdot\bm{\alpha}^w_t$ and $\mathbf{b}_t=\mathbf{b}_b\cdot\bm{\alpha}^b_t$.
	Different from the original structure in video understanding, online TAdaConv processes one frame at a time. Hence, it only considers the temporal context in the past just like tracking in the real world.
	Specifically, we keep a temporal context queue $\mathbf{\hat{X}}\in\mathbb{R}^{L\times C} $of $L$ frame descriptors $\mathbf{\hat{X}}_t\in\mathbb{R}^{C}$ including that of the current frame: 
	\begin{equation}
		\mathbf{\hat{X}} = \text{Cat}(\mathbf{\hat{X}}_t, \mathbf{\hat{X}}_{t-1},...,\mathbf{\hat{X}}_{t-L+1})\ ,
	\end{equation}
	\noindent where \text{Cat} represents the concatenation and the frame descriptor is obtained by a global average pooling (GAP) over the feature of the each coming frame, \textit{i.e.,} $\mathbf{\hat{X}}_t=\text{GAP}(\mathbf{X}_t)$.
	For the generation of calibration factors $\bm{\alpha}^w_t$ and $\bm{\alpha}^b_t$, we perform two convolutions over the temporal context queue $\mathbf{\hat{X}}$ with a kernel size of $L$, \textit{i.e.},
	$\bm{\alpha}^w_t =\mathcal{F}_\text{w}(\mathbf{\hat{X}}) + 1,~\bm{\alpha}^b_t=\mathcal{F}_\text{b}(\mathbf{\hat{X}}) + 1
	$, where $\mathcal{F}_i$ denotes the convolution operation. Besides, the weights of $\mathcal{F}$ are initialized to zero so that at the initialization, $\mathbf{W}_t=\mathbf{W}_b$ and $\mathbf{b}_t=\mathbf{b}_b$. For $t\leq L-1$, where there is not enough previous frames, we fill that with the descriptor of the first frame $\mathbf{\hat{X}}_1$. 
	Given our backbone $\varphi_{tada}$ that considers the temporal contexts in the feature extraction process, the similarity map $\textbf{R}_t$ for the $t$-th frame can be obtained as:
	\begin{equation}\label{11}
		\textbf{R}_t=\varphi_{tada}(\textbf{Z})\star\varphi_{tada}(\textbf{X}_t)~,
	\end{equation}
	\noindent where $\mathbf{Z}$ denotes the template and $\star$ represents the depth-wise correlation~\cite{8954116}. After that, $\textbf{F}_t$ can be obtained by a convolution layer, \textit{i.e.}, $\textbf{F}_t=\mathcal{F}(\textbf{R}_t)$.
	
	\Remark To the best of our knowledge, our online TAdaCNN is the first to integrate temporal contexts in the feature extraction process in the tracking task.

	\subsection{Similarity Refinement with AT-Trans}\label{tf}
	
	Besides considering temporal contexts in the feature extraction process, in this work, we also propose an AT-Trans for refining the similarity map $\mathbf{F}_t$ according to the temporal contexts.
	Specifically, our AT-Trans has an encoder-decoder structure, where the encoder aims to integrate temporal knowledge and the decoder focuses on similarity refinement.
	In this section, we first revisit the multi-head attention~\cite{aaat} before describing the details of our AT-Trans.
	
	\noindent\textbf{Multi-head attention.} 
	As a fundamental component of the transformer, multi-head attention is formulated as follows:
	\begin{equation}\label{e1}
		\begin{aligned}
			&\mathrm{MultiHead}(\mathbf{Q},\mathbf{K},\mathbf{V})=\Big(\mathrm{Cat}(\textbf{H}_{att}^1,...,\textbf{H}_{att}^N)\Big)\mathbf{W}\\
			&\textbf{H}_{att}^{n}=\mathrm{Attention}(\mathbf{Q}\mathbf{W}^n_q,\mathbf{K}\mathbf{W}_k^n,\mathbf{V}\mathbf{W}_v^n)\\
			&\mathrm{Attention}(\mathbf{Q},\mathbf{K},\mathbf{V})=\mathrm{Softmax}(\mathbf{Q}\mathbf{K}^{\mathrm{T}}/\sqrt{d})\mathbf{V}\\
		\end{aligned}
		~ ,
	\end{equation}
	where $\sqrt{d}$ is the scaling factor while $\mathbf{W}\in~\mathbb{R}^{C_i\times C_i}$, $\mathbf{W}^n_q\in~\mathbb{R}^{C_i\times C_h}$, $\mathbf{W}^n_k\in~\mathbb{R}^{C_i\times C_h}$, and $\mathbf{W}^n_v\in~\mathbb{R}^{C_i\times C_h}$ are learnable weights. In our AT-Trans, we employ multi-head attention with 6 heads, \textit{i.e.,} $N=6$ and $C_h$=$C_i/6$.

	Compared to CNN, Transformer can more effectively encode the global context information~\cite{aaat,dosovitskiy2021an}. Hence, to exploit the global temporal contexts more effectively, we propose a transformer-based temporal integration strategy to successively encode global contexts information. Moreover, most existing temporal-based methods generally store the input features for temporal modeling, inevitably introducing sensitive parameters and unnecessary computation. In this work, for eliminating unnecessary operations and sensitive parameters, we adopt an online update strategy for temporal knowledge.
	
	


	
\begin{figure}[t]
		\centering
		\includegraphics[width=1.0\linewidth]{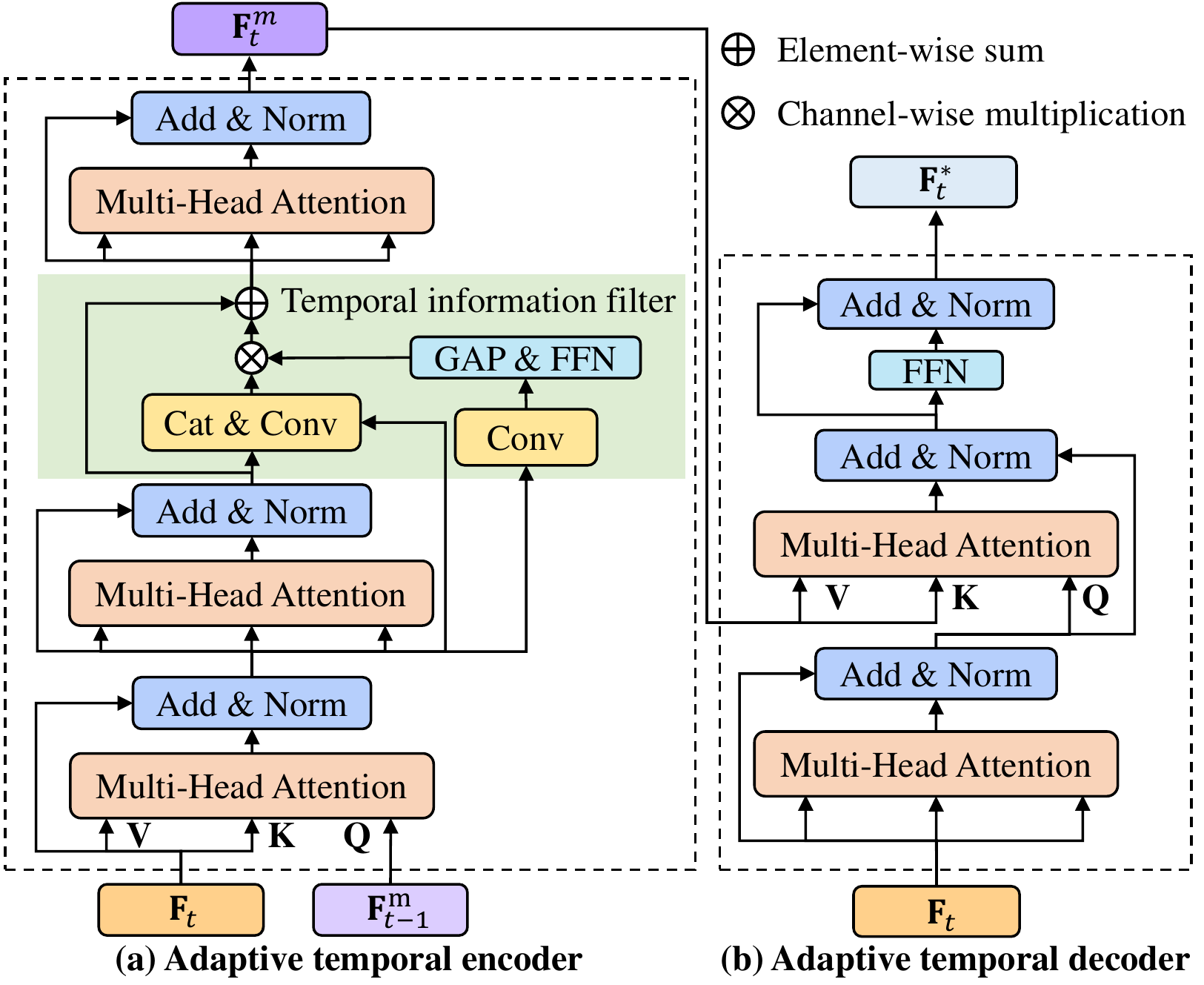}
		\vspace{-20pt}
		\caption{Structure of the adaptive temporal transformer. The left sub-window illustrates the adaptive temporal encoder to model the temporal knowledge. The right sub-window shows the component of the decoder. Best viewed in color.
		}
		\vspace{-10pt}
		\label{fig:work} 
	\end{figure}

	\noindent\textbf{Transformer encoder.} The encoder generates temporal prior knowledge by integrating the previous knowledge with current features. 
	Generally, we stack two multi-head attention layers before a temporal information filter is applied. 
	The final temporal prior knowledge for the current step is obtained by further attaching a multi-head attention layer to the filtered information. 
	The structure of the encoder is presented in Fig.~\ref{fig:work}(a).
	
	Given the previous temporal prior knowledge $\mathbf{F}_{t-1}^m$ and the current similarity map $\mathbf{F}_{t}$, there are two ways to integrate their information into the current prior knowledge $\mathbf{F}_{t}^m$, with respect to the selection of the query, key, and values. 
	One uses $\mathbf{F}_{t-1}^m$ as the query and $\mathbf{F}_{t}$ as the value and key, while the other uses them in reverse.
	In our method, we adopt the former, as this essentially puts more emphasis on the current similarity map. 
	This is plausible as closer temporal information is more valuable than the previous one for representing the characteristics of the current object more accurately. 
	Empirical results in Sec.~\ref{Sec:abla} also validate the effectiveness of this choice. 
	Hence, we obtain the output of the stacked multi-head attention layer in $t$-th frame $\mathbf{F}^{2}_t$ by:
	\begin{equation}
		\begin{aligned}
			& \mathbf{F}^{1}_t=\mathrm{Norm}(\mathbf{F}_{t}+\mathrm{MultiHead}(\mathbf{F}_{t-1}^m,\mathbf{F}_{t},\mathbf{F}_{t})) \\
			& \mathbf{F}^{2}_t=\mathrm{Norm}(\mathbf{F}^{1}_{t}+\mathrm{MultiHead}(\mathbf{F}^{1}_{t},\mathbf{F}^{1}_{t},\mathbf{F}^{1}_{t}))
		\end{aligned} \ ,
	\end{equation}
	where $\mathrm{Norm}$ represents the layer normalization. 

	Since aerial tracking may frequently encounter less useful contexts caused by motion blur or occlusion, some unwanted contexts may be included if we pass along the complete temporal information without any filtering. To eliminate the unwanted information, a neat temporal information filter is generated by attaching a feed-forward network $\mathrm{FFN}$ to the global descriptor of $\mathbf{F}^{1}_t$ obtained by global average pooling $\mathrm{GAP}$, \textit{i.e.,} $\bm{\alpha}=\mathrm{FFN}(\mathrm{GAP}(\mathcal{F}(\mathbf{F}^{1}_t)))$. The filtered information $\mathbf{F}^{f}_t$ is obtained by:
	\begin{equation}
		\mathbf{F}^{f}_t=\mathbf{F}^{2}_t+\mathcal{F}(\mathrm{Cat}(\mathbf{F}^{2}_t,\mathbf{F}^{1}_t))*\bm{\alpha}
		~ ,
	\end{equation}
	where $\mathcal{F}$ denotes a convolution layer. 
	With this, the temporal knowledge of \textit{t}-th frame, $\mathbf{F}^{m}_t$ can be obtained as follows:
	\begin{equation}
		\mathbf{F}^{m}_t=\mathrm{Norm}(\mathbf{F}^{f}_t+\mathrm{MultiHead}(\mathbf{F}^{f}_t,\mathbf{F}^{f}_t,\mathbf{F}^{f}_t))
		~ .
	\end{equation}

	Hence, for each frame, we update the temporal knowledge rather than saving all of them. This makes the memory occupancy of the temporal prior knowledge fixed during the whole tracking process, which makes TCTrack memory-efficient compared to approaches that require saving all the intermediate temporal information. Overall, owing to this strategy as well as the temporal filter and the multi-head attention, our AT-Trans adaptively encodes the temporal prior in a memory-efficient way.
	
	For the first frame in a tracking sequence, since the characteristics of different targets are distinct, using a unified initialization for the initial temporal prior $\mathbf{F}_{0}^m$ would be unreasonable. 
	Observing that the similarity map in the first frame essentially represents the semantic features of the target object in an effective way, we set the initial temporal prior by a convolution over the initial similarity map $\mathbf{F}_{0}$, \textit{i.e.,} $\mathbf{F}_{0}^m=\mathcal{F}_{init}(\mathbf{R}_{1})$. We also empirically show our initialization is better in Sec.~\ref{Sec:abla}.



	\begin{figure}[t]
		\centering

		\includegraphics[width=0.47\textwidth]{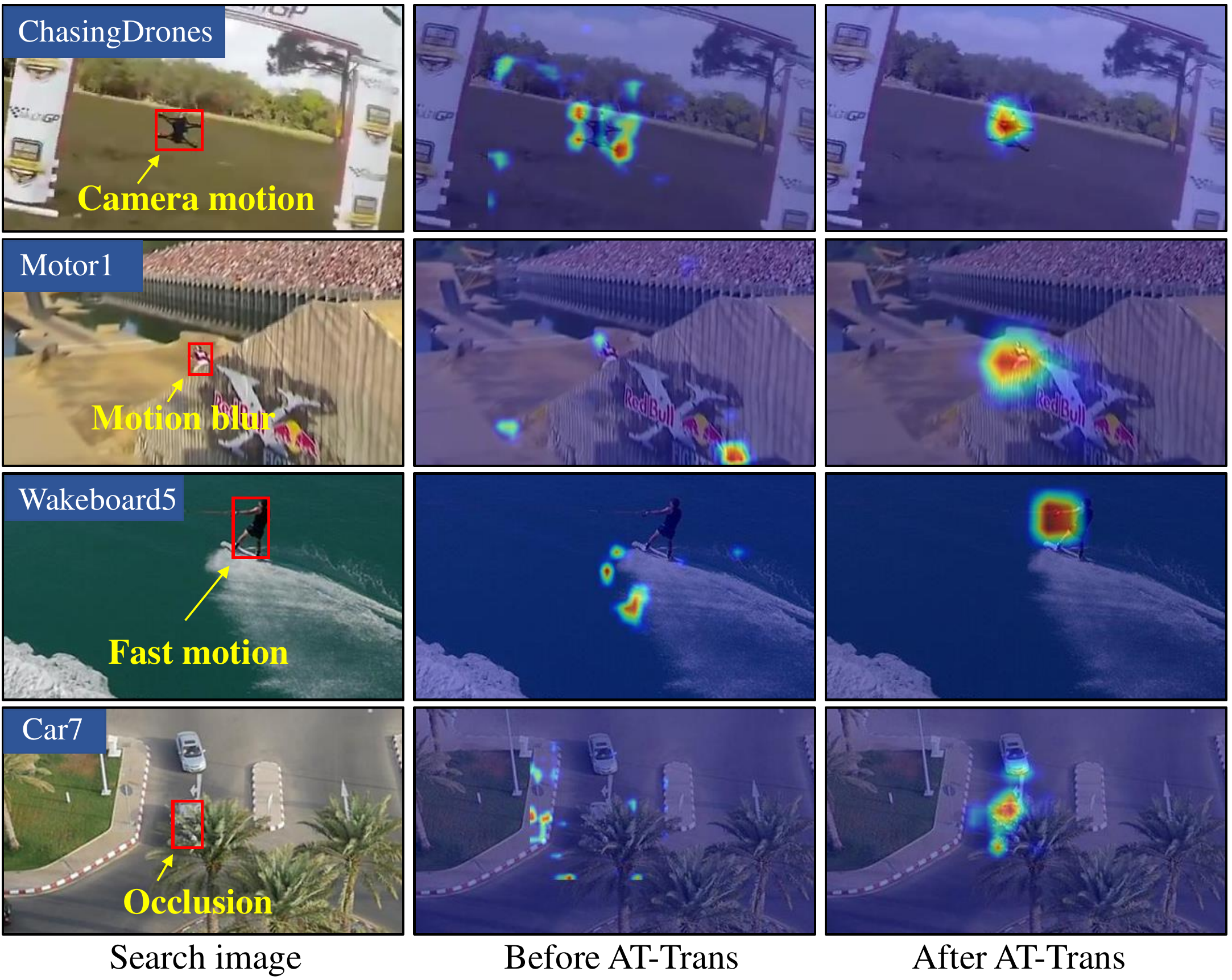}
		\vspace{-10pt}
		\caption{Comparison between similarity maps before refinement (second column) and after (third column) refinement.}
		\label{fig:vis}
		\vspace{-16pt}
	\end{figure}

	\noindent\textbf{Transformer decoder.} 
	According to the temporal prior knowledge $\mathbf{F}^{m}_t$, the decoder aims to refine the similarity map. 
	To better explore the interrelations between temporal knowledge and current spatial features $\mathbf{F}_t$, we adopt two multi-head attention layers with feed-forward before output. 
	Its structure is presented in Fig.~\ref{fig:work}(b).
	By generating the attention map, the valid information in the temporal knowledge $\mathbf{F}^{m}_t$ can be extracted for refining the similarity map $\mathbf{F}_t$ to obtain the final output $\mathbf{F}^*_t$:
	\begin{equation}
		\begin{split}
			&\mathbf{F}^3_t=\mathrm{Norm}(\mathbf{F}_t+\mathrm{MultiHead}(\mathbf{F}_t,\mathbf{F}_t,\mathbf{F}_t))\\
			&\mathbf{F}^4_t=\mathrm{Norm}(\mathbf{F}^3_t+\mathrm{MultiHead}(\mathbf{F}^3_t,\mathbf{F}^m_t,\mathbf{F}^m_t))\\
			&\mathbf{F}^{*}_t=\mathrm{Norm}(\mathbf{F}^{4}_t+\mathrm{FFN}(\mathbf{F}^{4}_t))\\
		\end{split}
		~ .
	\end{equation}
	
	Relying on the encoder-decoder structure of AT-Trans, the temporal contexts are effectively exploited to refine the similarity maps for boosting robustness and accuracy. The comparison of similarity maps in Fig.~\ref{fig:vis} shows the effectiveness of the similarity map refinement, especially where camera motion, severe motion, and occlusion exist.

	\Remark To the best of our knowledge, AT-Trans is the first attempt to use temporal contexts for similarity maps.
	
	\begin{figure*}[t]
		\centering	
		\includegraphics[width=0.325\textwidth]{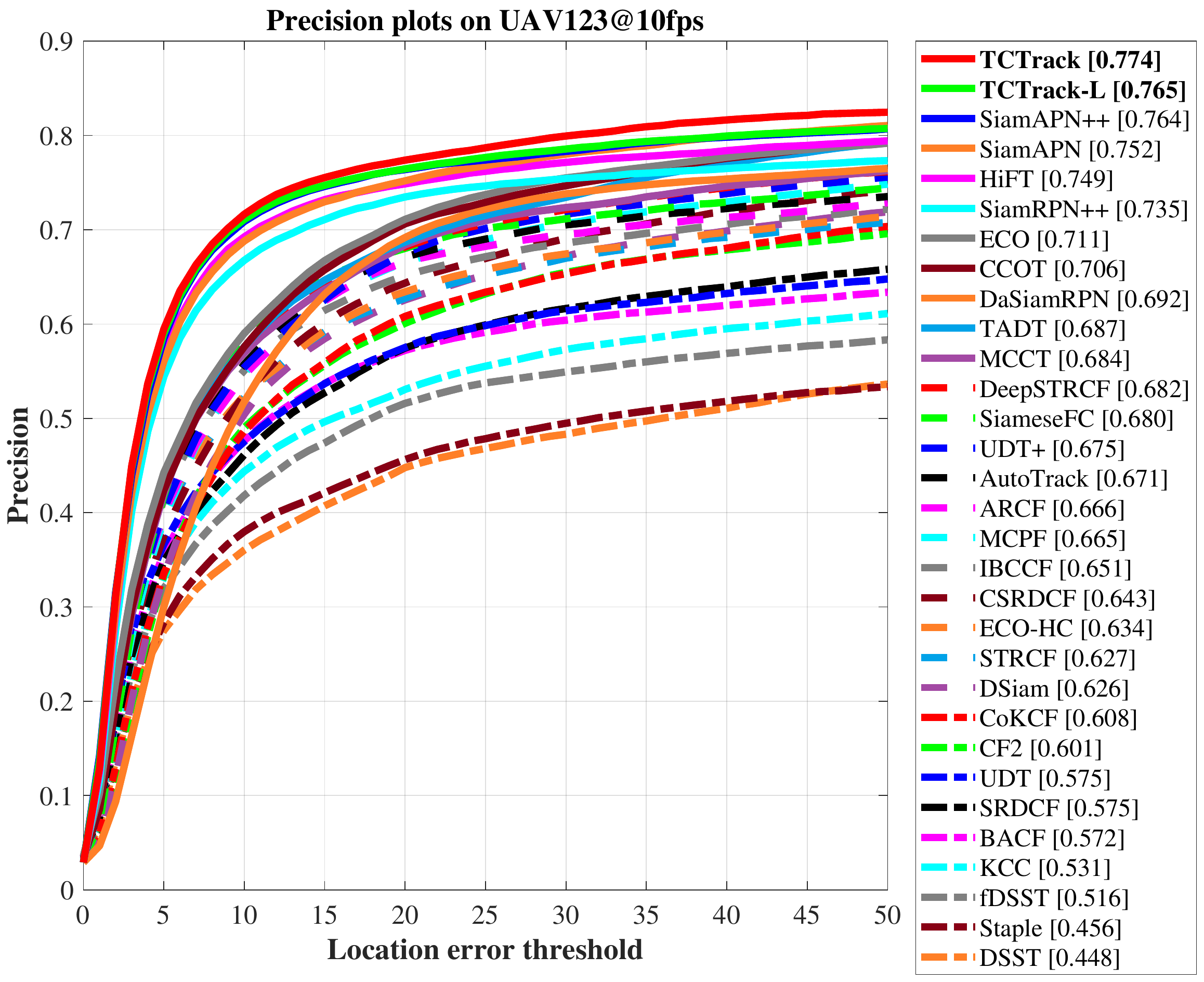}
		\includegraphics[width=0.325\textwidth]{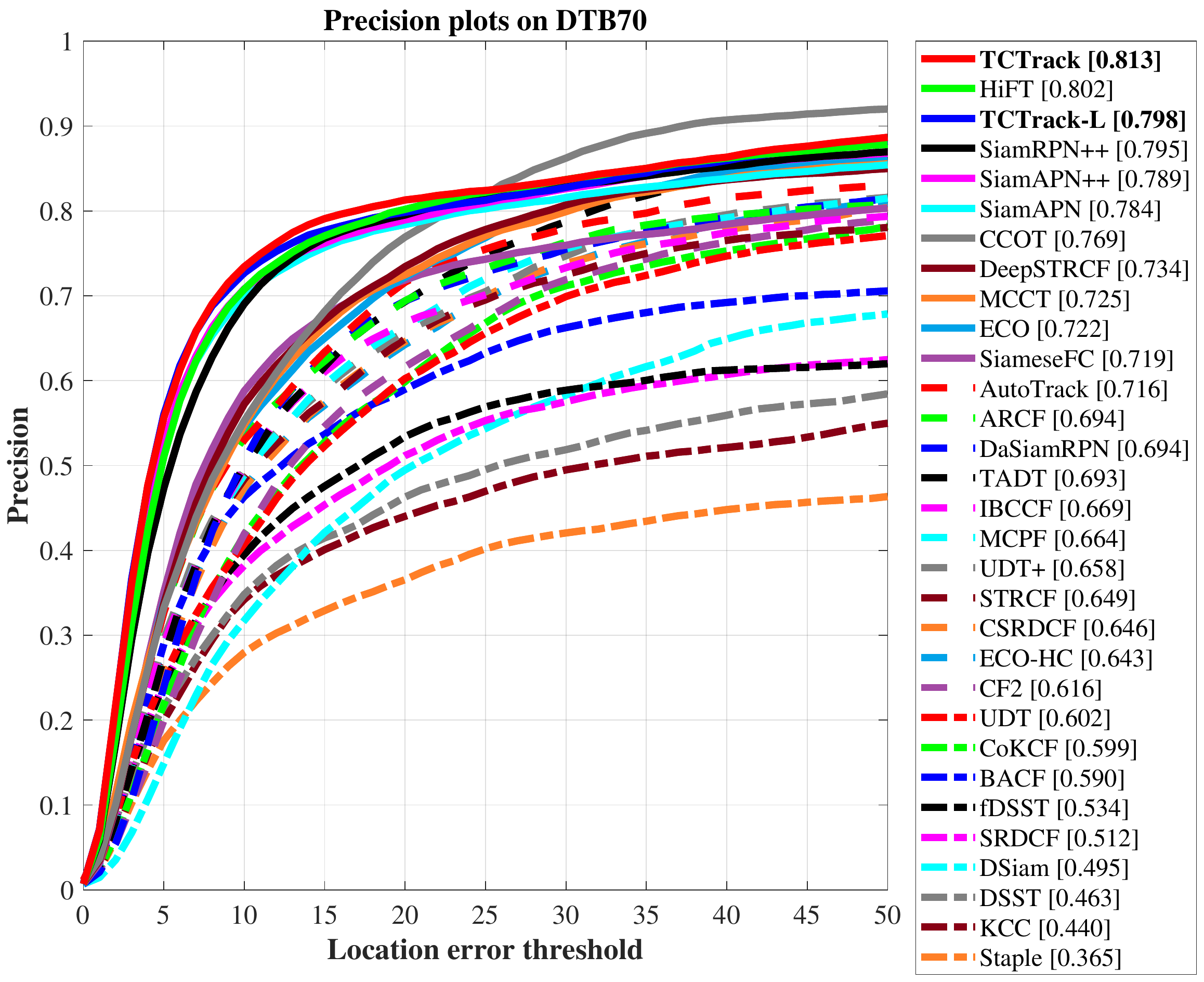}
		\includegraphics[width=0.325\textwidth]{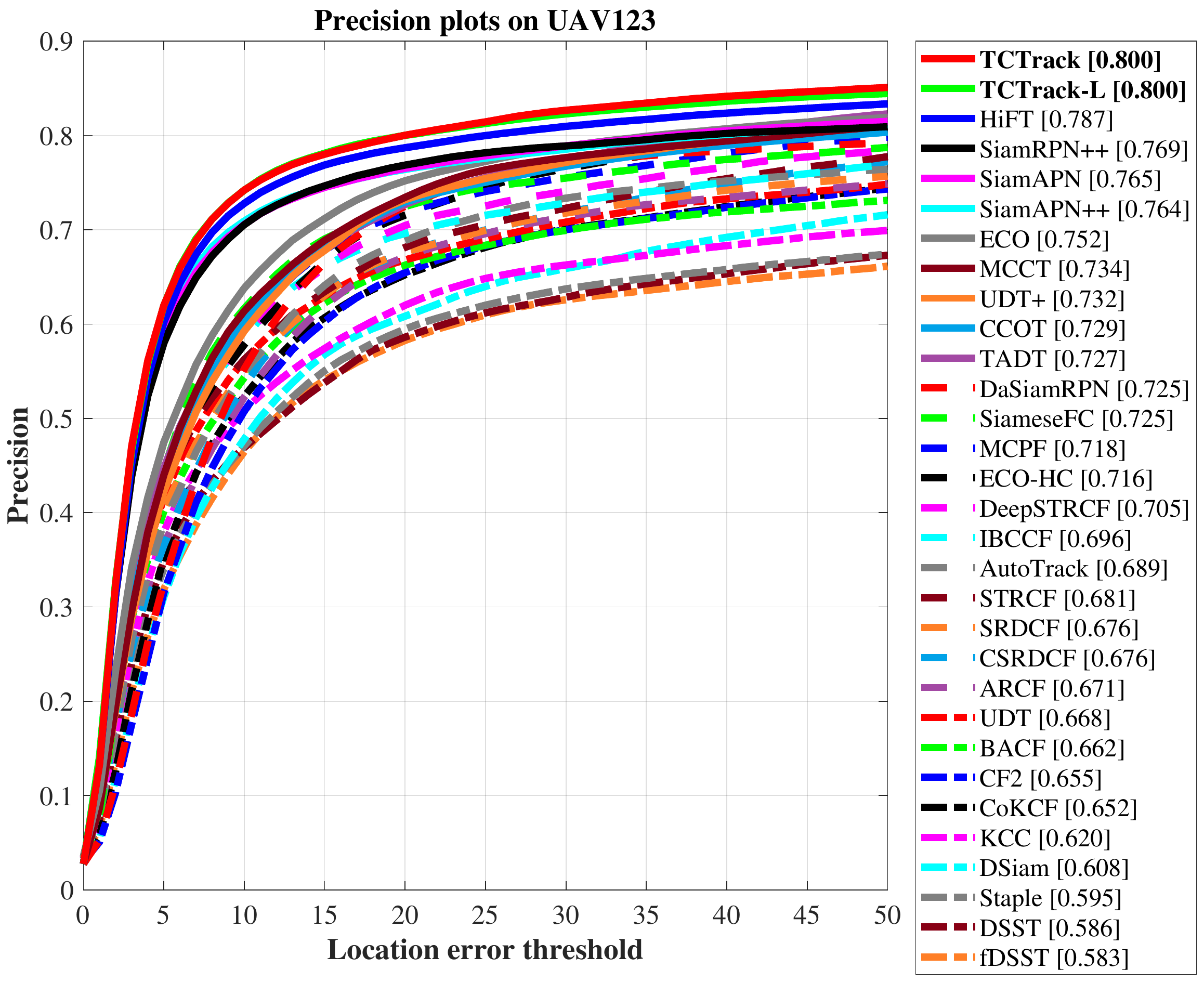}
		
	\end{figure*}
	\begin{figure*}[t]
		\centering	
		\vspace{-10pt}
		\includegraphics[width=0.325\textwidth]{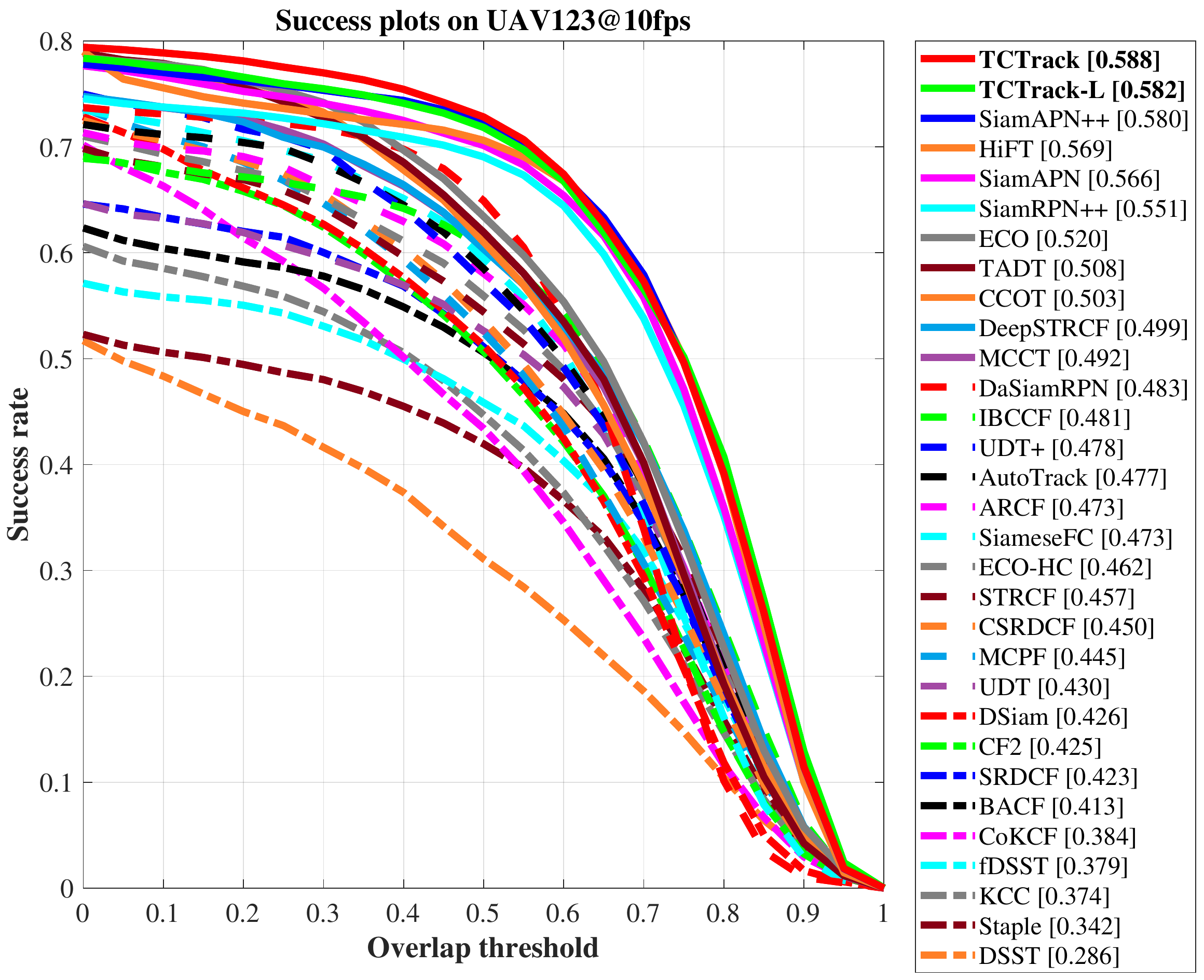}
		\includegraphics[width=0.325\textwidth]{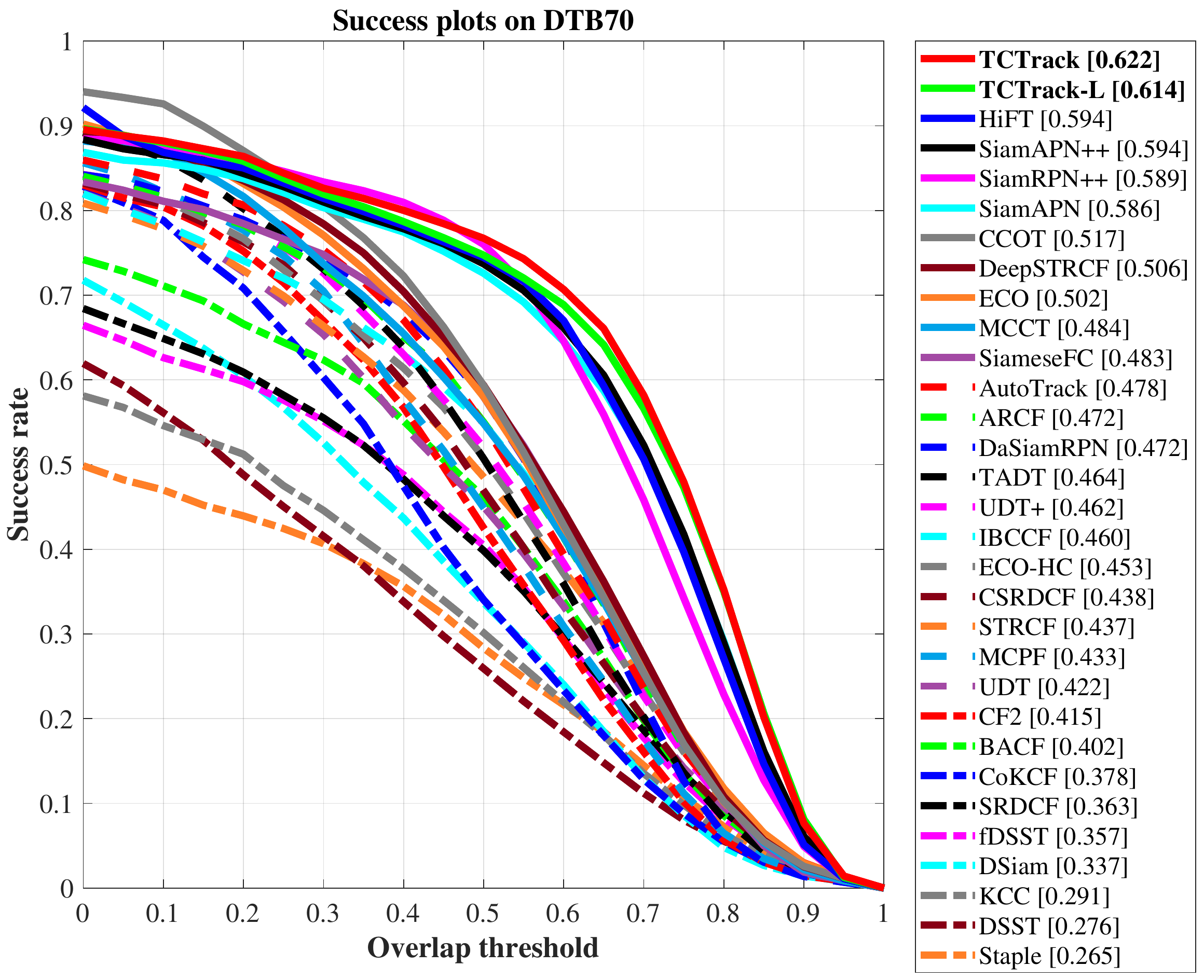}
		\includegraphics[width=0.325\textwidth]{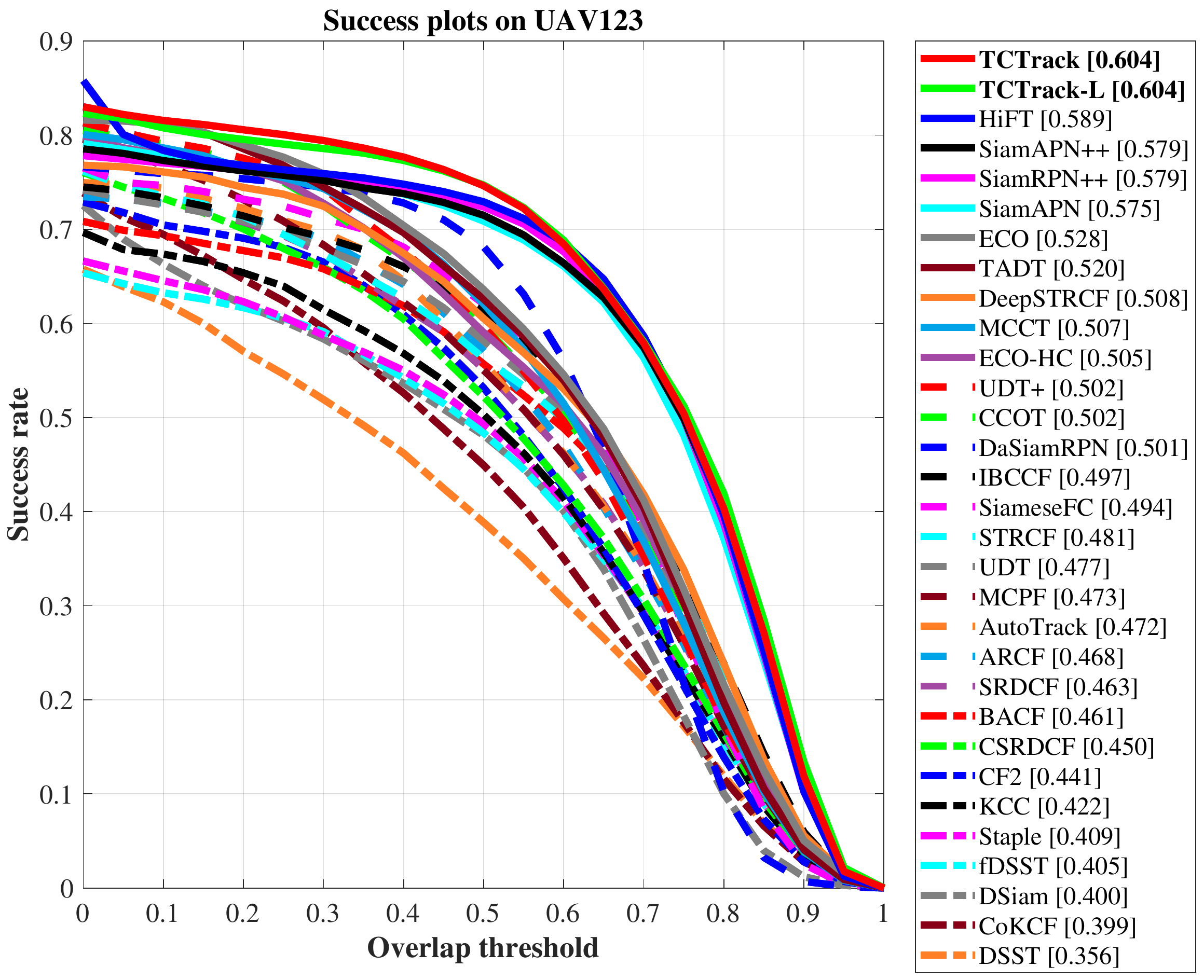}
		
		\vspace{-10pt}
		\caption{Overall performance of all trackers on three well-known aerial tracking benchmarks. 
			Our tracker achieves superior performance against other SOTA trackers. \Trackname{}-L represents the tracker with AT-Trans while the \Trackname{} denotes the full version of our framework.}
		\label{fig:ov}
		\vspace{-15pt}
	\end{figure*}
	

	\section{Experiments}\label{sec:Evaluation}
	
	Our framework is evaluated on four public authoritative benchmarks and tested on real-world aerial tracking conditions. 
	In this section, our method is comprehensively evaluated on four well-known aerial tracking benchmarks, \textit{i.e.}, UAV123~\cite{Mueller2016ECCV}, UAVTrack112\_L~\cite{9477413}, UAV123@10fps~\cite{Mueller2016ECCV}, and DTB70~\cite{li2017visual}. 
	51 existing top trackers are included for a thorough comparison, where their results are obtained by running the official codes with their corresponding hyper-parameters. For a clearer comparison, we divide them into two groups, \textbf{(i)} light-weight trackers~\cite{8954116,cao2021iccv,fu2020siamese,cao2021siamapn,8100216,8578607,Wang_2019_Unsupervised,danelljan2016beyond,Xin2019CVPR,zhu2018distractor,siamfc,zhang2017multi,8578613,li2017integrating,Li_2020_CVPR,SRDCF,lukezic2017discriminative,Huang2019ICCV,kiani2017learning,ma2015hierarchical,zhang2017robust,wang2018kernel,dsiam,bertinetto2016staple,danelljan2016discriminative,danelljan2014accurate} and \textbf{(ii)} deep trackers~\cite{mayer2021learning,zhang2020ocean,Guo_2021_CVPR,updatenet,siamdw,8953466,xu2020siamfc++,8954116,Chen_2021_CVPR,chen2020siamese,wang2021transformer,lukezic2020d3s,fu2021stmtrack,9157720,9157124,sosnovik2021scale,9010649,8953931}.

	
	%
	%

	%
	\begin{figure*}[t]
		\centering	
		\vspace{-10pt}
		\includegraphics[width=0.24\textwidth]{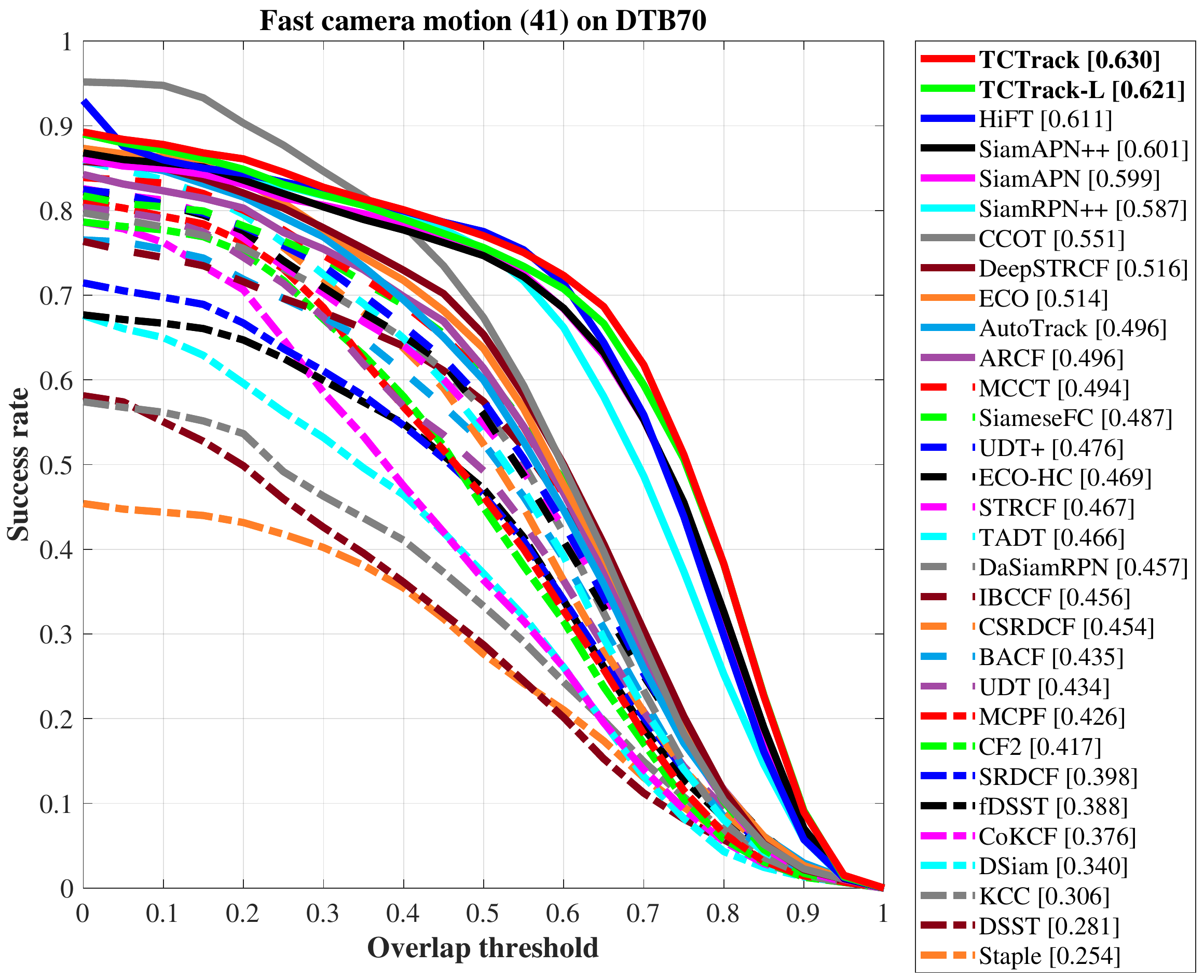}
		\includegraphics[width=0.24\textwidth]{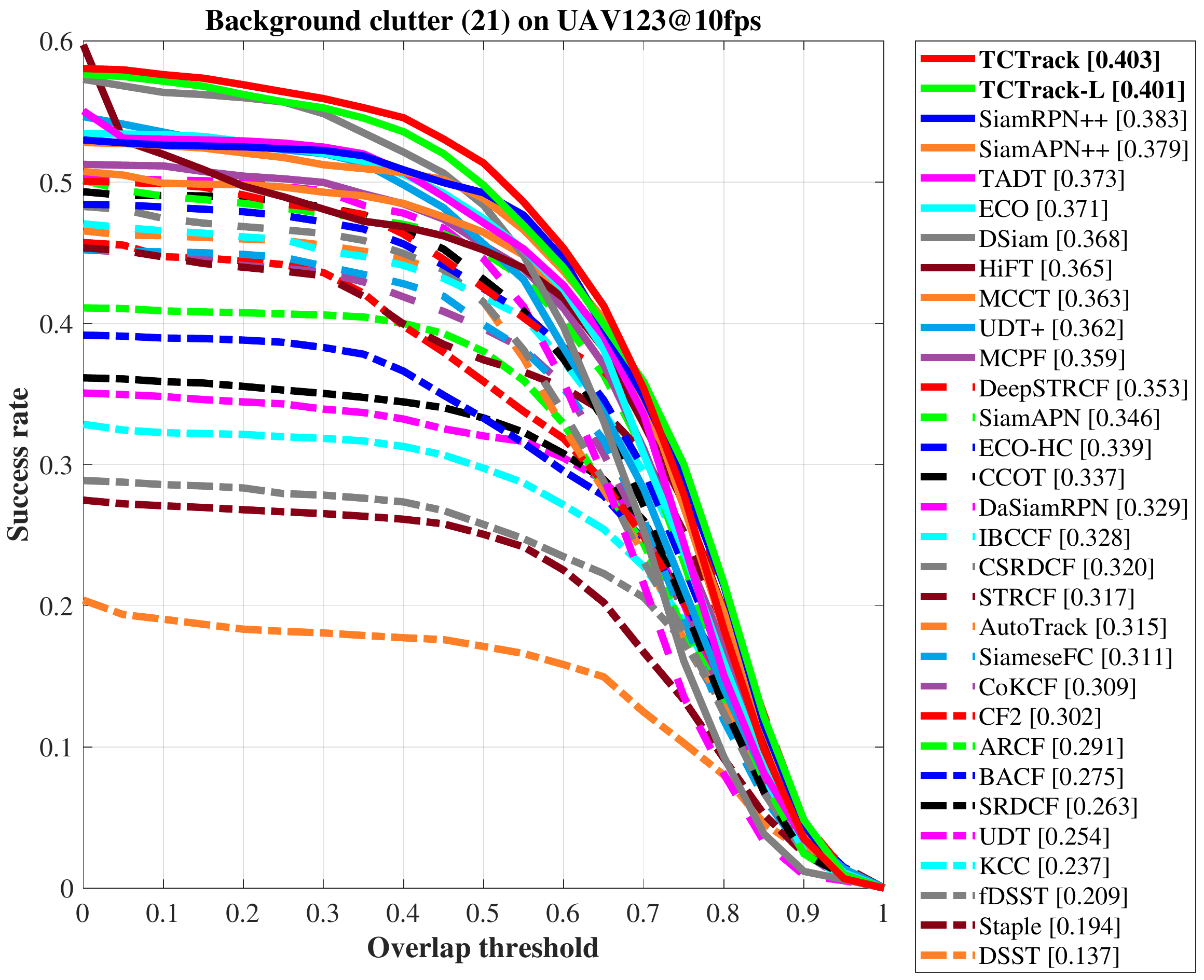}
		\includegraphics[width=0.24\textwidth]{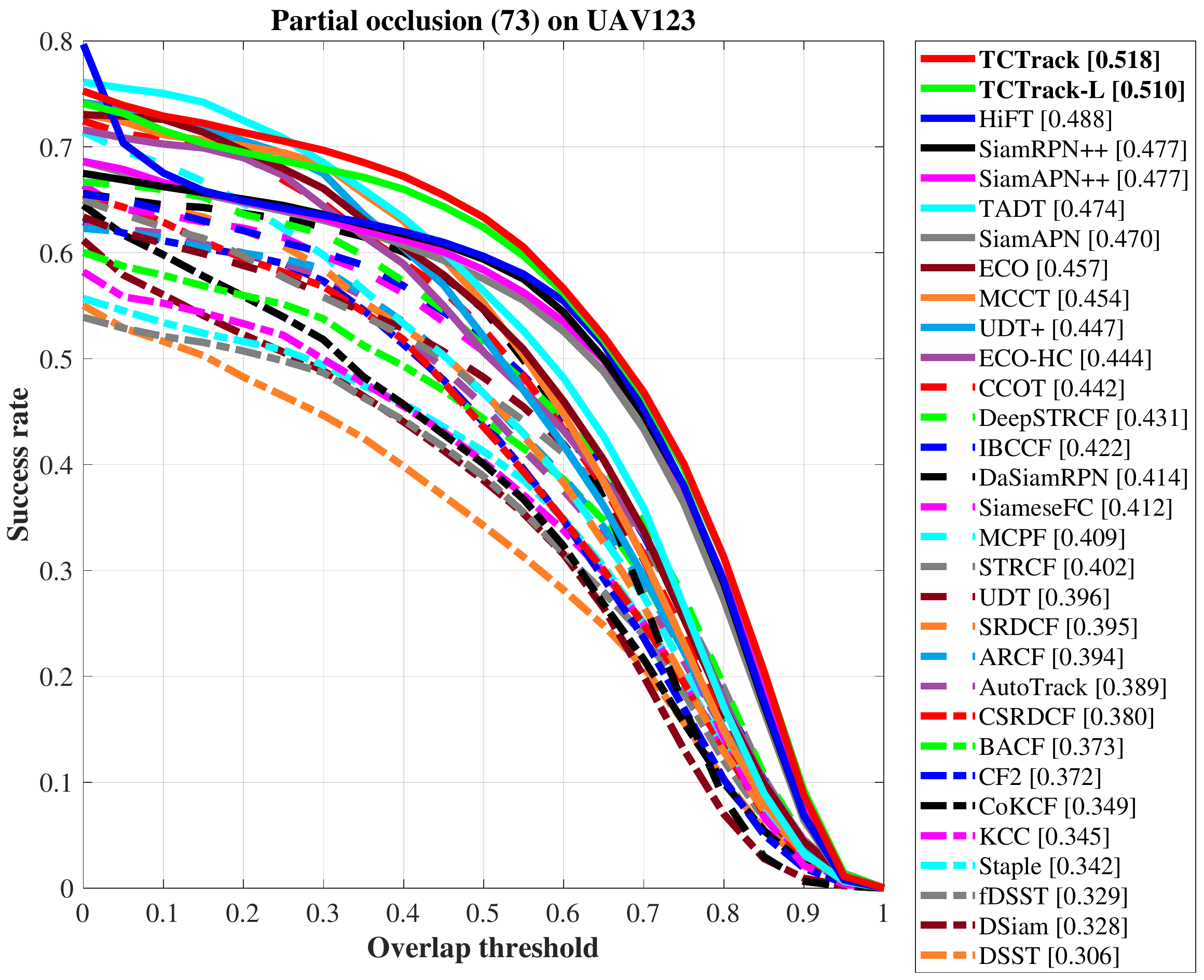}
		\includegraphics[width=0.24\textwidth]{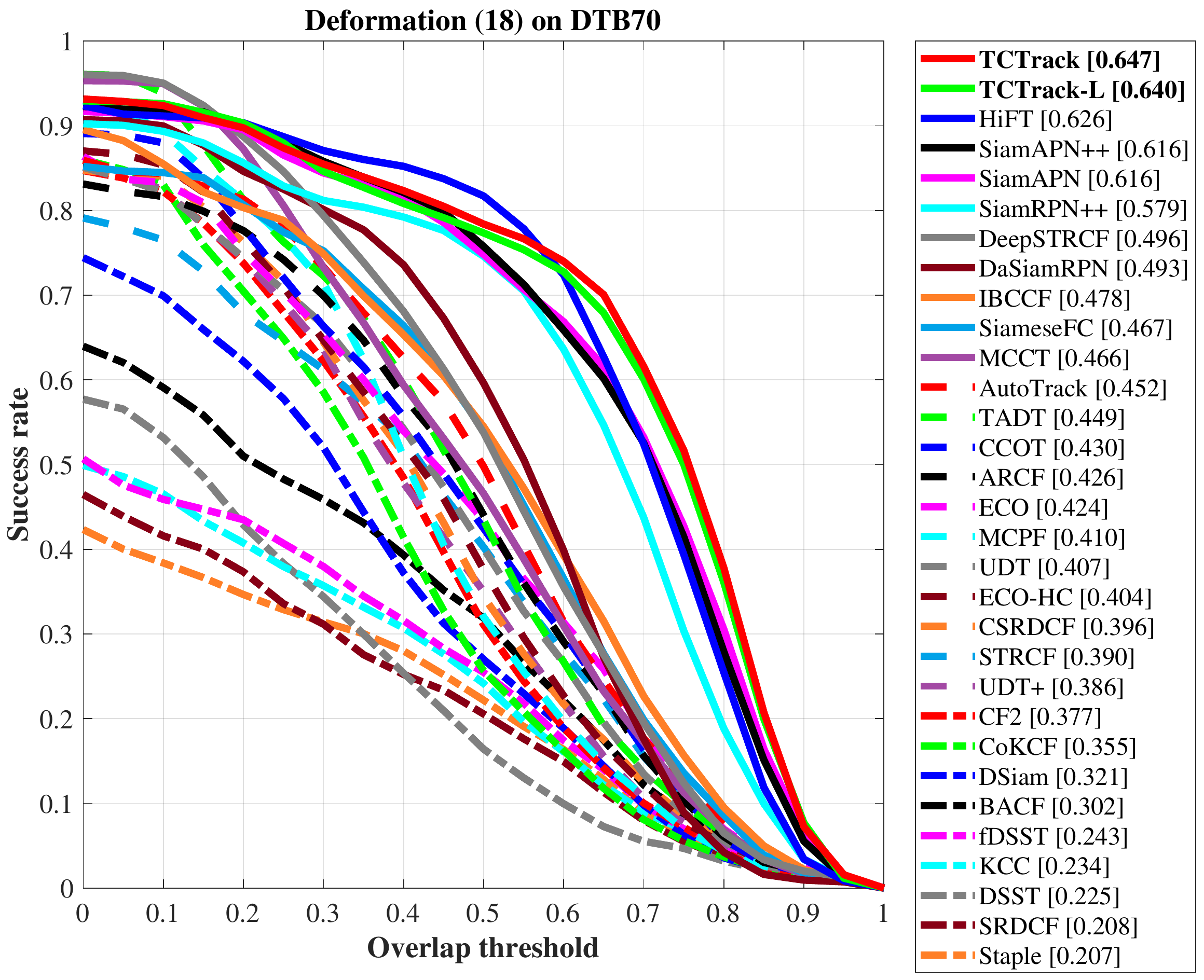}
		\vspace{-5pt}
		\caption{Attribute-based evaluation of all trackers on three well-known aerial tracking benchmarks. Our temporal tracker can maintain promising performance under severe motion, occlusion, and deformation. More results are shown in the supplementary material.}\label{fig:att}
		\vspace{-5pt}
		
	\end{figure*}

	\subsection{Implementation Details}
	
	We use AlexNet as the backbone of our tracker, as efficiency is essential for aerial tracking. As shown in Table~\ref{tab:back}, the comparison in inference time of different popular backbones on the NVIDIA Jetson AGX Xavier platform has shown that AlexNet has the lowest latency, while the recent developments in mobile networks~\cite{iandola2016squeezenet,8578572,8578814} suffer from high memory access cost (MAC). For initialization, we use ImageNet pre-trained model for AlexNet and use the same initialization for online TAdaConv as in~\cite{russakovsky2015imagenet}. The AT-Trans in our TCTrack is randomly initialized.


	\begin{table}[t]
		\footnotesize
		\setlength{\tabcolsep}{4mm}
		\centering
		\caption{Comparison of inference time and parameters on NVIDIA Jetson AGX Xavier. Here, we use 287$\times$287$\times$3 as the input image and only evaluate the inference time of the CNN. 
		}
		\vspace{-8pt}
		\centering
		{
			\begin{tabular}{l|c|c }
				\toprule[2pt]
				
				{\textbf{Backbone}}&Inference time &Parameters\\
				\midrule
				{AlexNet~\cite{krizhevsky2012imagenet}}&3.4ms&2.47M\\
				{VGG11~\cite{simonyan2014very}}&3.7ms&9.22M\\
				{ResNet18~\cite{7780459}}&10.1ms&11.2M\\
				{MobileNet\_v2~\cite{8578572}}&13.7ms&2.2M\\
				{EfficientNet~\cite{tan2019efficientnet}}&27.4ms&39.4K\\
				{SqueezeNet1\_0~\cite{iandola2016squeezenet}}&8.8ms&735.42K\\
				{ShuffleNet\_v2\_x0.5~\cite{8578814}}&16.6ms&341.8K\\ 
				
				\bottomrule[2pt]
			\end{tabular}%
		}
		\vspace{-20pt}
		\label{tab:back}%
	\end{table}%

	We train our tracker with the videos whose length are 4 from VID~\cite{russakovsky2015imagenet}, Lasot~\cite{fan2019lasot}, and GOT-10K~\cite{huang2019got}. We train TCTrack for a total of 100 epochs on two NVIDIA TITAN RTX GPUs. For the first 10 epochs, the parameters of the backbone are frozen, following~\cite{8954116}.
	The rest of the training process employs a learning rate decreasing from $0.005$ to $0.0005$ in log space. 
	SGD is employed as the optimizer with a momentum of $0.9$, where the mini-batch size is $124$ pairs.
	The input sizes of the template and the search area are $127^2$ and $287^2$ respectively. The proposed online TAdaConv is used in the replacement of the last two convolutional layers.

	\Remark For more detailed information about the evaluation criteria and loss function, please refer to the supplementary material.
	


	\subsection{Comparison with Light-Weight Trackers}~\label{over}
	In this subsection, TCTrack is compared with 29 existing efficient trackers on the standard aerial tracking benchmarks.
	For Siamese-based methods, we evaluate them with the same backbone as ours for a fair comparison.
	


	\noindent\textbf{UAV123.} UAV123~\cite{Mueller2016ECCV} is a large-scale aerial tracking benchmark involving 123 challenging sequences with more than 112\textit{K} frames. Performance evaluation on UAV123 can verify the tracking performance in most commonly aerial tracking conditions. As shown in Fig.~\ref{fig:ov}, our \Trackname{} outperforms HiFT and SiamRPN++ in AUC (3\%) and (4.3\%).

	\noindent\textbf{DTB70.} DTB70~\cite{li2017visual} includes 70 severe motion scenarios in various challenging scenes. For evaluating the effectiveness of our method in handling motion, we adopt this benchmark to prove the robustness of \Trackname. Our tracker ranks 1st with an improvement of 5\% in AUC against the other best tracker illustrated in Fig.~\ref{fig:ov}.

	\noindent\textbf{UAV123@10fps.} Adopting an image rate of 10 FPS, the motion and variation are more abrupt and severe in UAV123@10fps~\cite{Mueller2016ECCV}, thereby significantly raising the difficulty of tracking. From the comparison with our other SOTA trackers, we can clearly see that our tracker maintains superior robustness and exceeds the second-best tracker in terms of success and precision rate.

	\noindent\textbf{Attribute-based performance.}
	\label{subsec:attr}
	In aerial tracking conditions, the severe motion of UAVs will increase the difficulty of tracking. To fully analyze the robustness of our tracker in specific challenges such as fast motion, camera motion, occlusion, deformation, \textit{etc}, attribute-based comparisons are conducted. The comparison between other SOTA trackers presented in Fig.~\ref{fig:att} proves the robustness of our framework in several challenging conditions. Since our tracker can accumulate the consecutive temporal knowledge from 1st frame to the current frame, our tracker can learn the historical location of the object. Therefore, our tracker achieves superior performance in occlusion and fast-motion scenes. Furthermore, benefiting from our content-adaptive temporal knowledge and online TAdaConv, TCTrack can handle the negative influence introduced by the environment.

	\begin{table}[t]
		\caption{Overall performance on UAVTrack112\_L. The best three
			performances are respectively highlighted with \textcolor[rgb]{ 1,  0,  0}{\textbf{red}}, \textcolor[rgb]{ 0,  1,  0}{\textbf{green}}, and \textcolor[rgb]{ 0,  0,  1}{\textbf{blue}} colors.}
		\vspace{-10pt}
		\centering
		\renewcommand\tabcolsep{8pt}
		\resizebox{0.99\linewidth}{!}{
			\begin{tabular}{lcc||lcc}
				\hline
				\hline
				Trackers & Succ.  & Prec.  & Trackers & Succ.  & Prec. \\
				\hline
				AutoTrack~\cite{Li_2020_CVPR} & 0.405  & 0.675  & C-COT~\cite{danelljan2016beyond} & 0.422  & 0.691 \\
				ARCF~\cite{Huang2019ICCV} & 0.399  & 0.640  & 
				UDT+~\cite{Wang_2019_Unsupervised} & 0.405  & 0.637  \\
				STRCF~\cite{8578613} & 0.360  & 0.609  & ECO~\cite{8100216} & 0.436  & 0.684 \\
				UDT~\cite{Wang_2019_Unsupervised} & 0.388  & 0.620  & TADT~\cite{Xin2019CVPR} & 0.462  & 0.712 \\
				SRDCF~\cite{SRDCF} & 0.320  & 0.508  &SiameseFC~\cite{siamfc} & 0.452  & 0.690 \\
				CoKCF~\cite{zhang2017robust} & 0.283  & 0.520  & DaSiamRPN~\cite{zhu2018distractor} & 0.479  & 0.729 \\
				
				BACF~\cite{kiani2017learning} & 0.358  & 0.593  & SiamAPN++~\cite{cao2021siamapn} & 0.537 & \textcolor[rgb]{ 0,  0,  1}{\textbf{0.735}} \\
				DSiam~\cite{dsiam} & 0.321 & 0.512 & SiamRPN++~\cite{8954116} & \textcolor[rgb]{ 0,  1,  0}{\textbf{0.559}}  & \textcolor[rgb]{ 0,  1,  0}{\textbf{0.773}}\\
				HiFT~\cite{cao2021iccv} & \textcolor[rgb]{ 0,  0,  1}{\textbf{0.551}}  & 0.734  & \textbf{\Trackname{}~ (ours)} & \textcolor[rgb]{ 1,  0,  0}{\textbf{0.582}} & \textcolor[rgb]{ 1,  0,  0}{\textbf{0.786}} \\
				\hline
				\hline
		\end{tabular}}
		\vspace{-20pt}
		\label{tab:20}
	\end{table}%

	\begin{table*}[t]
		\centering
		\scriptsize 
		\caption{Ablation study of different components of adaptive temporal transformer on UAV123~\cite{Mueller2016ECCV}. \texttt{TIF} denotes the temporal information filter in the AT-Trans (Fig.~\ref{fig:work}). \texttt{SF/MF} refer to single-frame (SF) training, \textit{i.e.,} the standard tracking-by-detection training method and our multi-frame (MF) training method. \texttt{CI/RI} refer to convolutional initialization and random initialization for temporal prior knowledge. Query denotes which feature map is used as the query in the adaptive temporal encoder in AT-Trans mentioned in Sec.~\ref{tf}. 
		}
		\vspace{-5pt}
		\renewcommand\tabcolsep{1pt}
		\begin{tabular}{lccc|cccccccc}
			\toprule
			& & & & \multicolumn{2}{c}{Camera Motion} &\multicolumn{2}{c}{Fast motion}&\multicolumn{2}{c}{Partial Occlusion} & \multicolumn{2}{c}{Overall} \\
			\midrule
			Model & Train & Init. & Query & Prec. & Succ. & Prec. & Succ.& Prec. & Succ.& Prec. & Succ. \\
			\midrule
			Transformer & SF & - & $\mathrm{F_{t-1}^m}$ &  $0.750$ & $0.549$&$0.712$ & $0.509$&$0.663$ & $0.458$& $0.750$ & $0.550$ \\
			Transformer+TIF & SF & - & $\mathrm{F_{t-1}^m}$ & $0.767_{2.3\%\uparrow}$ & $0.578_{5.3\%\uparrow}$ & $0.720_{1.1\%\uparrow}$ & $0.525_{3.1\%\uparrow}$ & $0.667_{0.6\%\uparrow}$ & $0.474_{3.5\%\uparrow}$  & $0.765_{2.0\%\uparrow}$ & $0.573_{4.2\%\uparrow}$ \\
			Transformer & MF & CI & $\mathrm{F_{t-1}^m}$ & $0.749_{2.4\%\downarrow}$ & $0.525_{7.6\%\downarrow}$ & $0.719_{2.4\%\downarrow}$ & $0.500_{7.6\%\downarrow}$ &  $0.639_{2.4\%\downarrow}$ & $0.415_{7.6\%\downarrow}$ & $0.732_{2.4\%\downarrow}$ & $0.508_{7.6\%\downarrow}$ \\
			Transformer+TIF & MF & RI & $\mathrm{F_{t-1}^m}$ & $0.779_{3.9\%\uparrow}$  & $0.592_{7.8\%\uparrow}$& $0.766_{7.6\%\uparrow}$  & $0.566_{11.2\%\uparrow}$& $0.670_{1.1\%\uparrow}$  & $0.483_{5.5\%\uparrow}$ & $0.772_{2.9\%\uparrow}$  & $0.586_{6.6\%\uparrow}$\\
			Transformer+TIF & MF & CI & $\mathrm{F_{t}}$ & $0.785_{4.7\%\uparrow}$ & $0.587_{6.9\%\uparrow}$& $0.726_{2.0\%\uparrow}$ & $0.528_{3.7\%\uparrow}$& $0.676_{2.0\%\uparrow}$ & $0.480_{4.8\%\uparrow}$ & $0.771_{2.8\%\uparrow}$ & $0.580_{5.5\%\uparrow}$ \\
			\textbf{Transformer+TIF} & \textbf{MF} & \textbf{CI} & $\bf{F_{t-1}^m}$ & $\bm{0.810_{8.0\%\uparrow}}$ & $\bm{0.615_{12.0\%\uparrow}}$& $\bm{0.793_{11.3\%\uparrow}}$ & $\bm{0.586_{15.1\%\uparrow}}$& $\bm{0.710_{7.1\%\uparrow}}$ & $\bm{0.510_{11.4\%\uparrow}}$& $\bm{0.800_{6.7\%\uparrow}}$ & $\bm{0.604_{9.8\%\uparrow}}$ \\ 
			\bottomrule
		\end{tabular}%
		\label{tab:ablati}%
		\vspace{-15pt}
	\end{table*}%

	\begin{table}[t]
		
		\centering
		\caption{Different sequence lengths for the online TAdaConv on UAV123~\cite{Mueller2016ECCV}. }
		\vspace{-5pt}
		\renewcommand\tabcolsep{4pt}
		\resizebox{1.0\linewidth}{!}{
			\begin{tabular}{lcc}
				\toprule
				
				Different Variations & Overall Precision  &Overall Success\\
				
				\midrule
				
				Transformer& $0.750$ & $0.550$ \\
				Transformer+TAdaConv (L=1)   & $0.749_{0.1\%\downarrow}$ & $0.561_{2.0\%\uparrow}$\\
				Transformer+TAdaConv (L=2)   & $0.774_{3.2\%\uparrow}$ & $0.573_{4.2\%\uparrow}$ \\
				\textbf{Transformer+TAdaConv (L=3)}   & $\bf{0.776_{3.5\%\uparrow}}$ & $\bf{0.580_{5.5\%\uparrow}}$ \\

				\bottomrule
		\end{tabular}}
		\vspace{-15pt}
		\label{tab:a}%
	\end{table}%

	\noindent\textbf{UAVTrack112\_L.} To validate the effectiveness of our framework in long-term tracking performance, we conduct the evaluations on UAVTrack112\_L~\cite{9477413}, which is the current biggest long-term aerial tracking benchmark including over 60k frames. Table~\ref{tab:20} reports the comparison of TCTrack and other SOTA trackers. Thanks to our comprehensive framework that fully exploits temporal contexts, \Trackname{} achieves superior performance against other trackers in terms of precision (0.786) and success rate (0.582).

	\subsection{Ablation Study}\label{Sec:abla}
	To verify the effectiveness of our framework, comprehensive ablation studies are presented in this subsection.

	%
	%
	%
	%
	%
	%
	%

	\noindent\textbf{Clarification of symbol.}
	In Table.~\ref{tab:ablati}, we denote our proposed transformer architecture without temporal information filter as \texttt{Transformer}. We analyze the influence caused by different models, training methods, initializations, and query selections. Furthermore, for ensuring the correctness of our experiments, all tracker adopts the same process (including training, parameter settings, \textit{etc.}) except for the studied module.

	\noindent\textbf{Analysis on AT-Trans.} \textbf{I}) Adding the consecutive temporal knowledge without filtering out the invalid information (third line) will confuse the tracker. Therefore, the tracking performance is impeded significantly. By adding our information filter in the tracking-by-detection framework, our module can also raise the performance by adaptively selecting valid contexts (second line). \textbf{II}) As we discussed before, using the unique information of the tracking object in the first frame to initiate the temporal knowledge is more appropriate than random initiation, especially in occlusion conditions (raising about 6\%). \textbf{III}) We also analyze the effect caused by the different queries. The results prove that refinement based on the current similarity map is more effective and suitable for raising performance, especially in motion scenarios (improved over 10\%). 
	
	Compared with \texttt{Transformer}, there is a significant improvement brought by our temporal knowledge encoded by AT-Trans (\textbf{9.8\%} in overall AUC and \textbf{6.7\%} in overall precision). Specifically, our tracker yields the best performance with an improvement of about \textbf{12.0\%} and \textbf{15.1\%} in handling the motion scenes. In the occlusion conditions, owing to the consecutive temporal contexts, our tracker can relocate the object via the previous information, thereby boosting the success rate by \textbf{11.4\%}.


	
	\noindent\textbf{Studies about the length of temporal sequences in TAdaConv.} As shown in Table.~\ref{tab:a}, when the image range of TAdaConv is increasing, the performance is raising. To introduce the temporal contexts effectively and efficiently, in this work, we adopt 3 as the length of sequences, \textit{i.e.}, L=3.

	\subsection{Comparison with Deep Trackers}\label{Sec:deep}
	
\begin{figure}[t]
		\centering
		\vspace{-10pt}
		\includegraphics[width=1\linewidth]{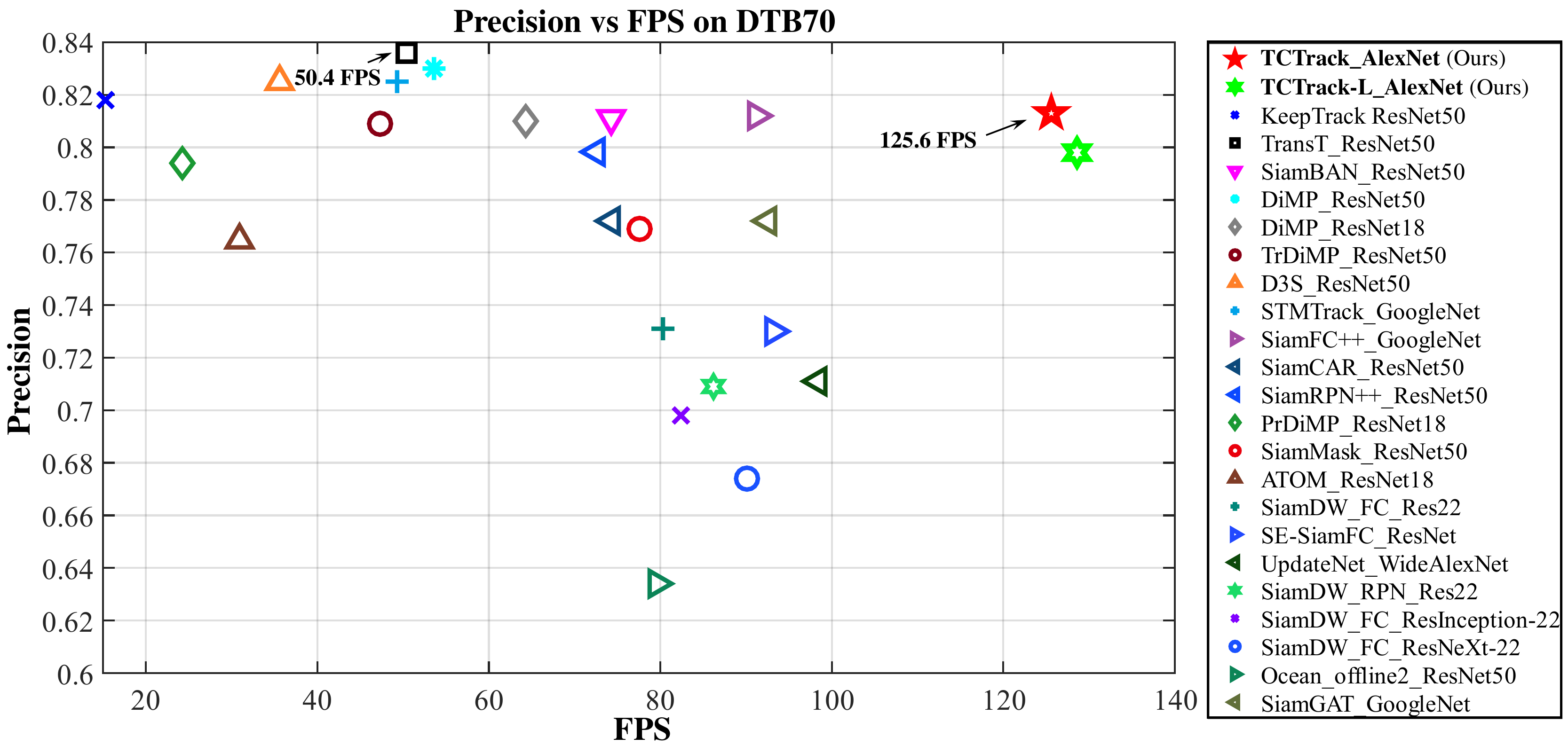}
		\vspace{-15pt}
		\caption{Comparisons to trackers with deeper backbones on DTB70.  Our tracker achieves competitive performance compared with other deeper trackers while possessing superior efficiency.}
		\vspace{-15pt}
		\label{fig:deeper_star}
	\end{figure}

	Our approach aims to introduce temporal information to raise the robustness and handle the challenges in aerial tracking. Therefore, to comprehensively illustrate our efficiency and performance against other SOTA trackers with deeper backbones, further comparisons are constructed including over 20 trackers on NVIDIA TITAN RTX. As illustrated in Fig.~\ref{fig:deeper_star}, although adopting the lightweight CNN as our backbone, \Trackname{} achieves competitive performance compared with the best tracker while running \textbf{2.49} times faster than the best tracker (TransT). Attribute to our content-adaptive and memory-efficient structure, our framework with temporal contexts can fill the performance margin caused by deeper backbones while maintaining the promising efficiency in aerial tracking conditions.
	

	
	\vspace{-5pt}
	\section{Real-world Tests}\label{sec:Real-world Tests}
	
	In this section, we implement our tracker on UAV to validate its practicability in real-world conditions. Specifically, NVIDIA Jetson AGX Xavier and Pixhawk\footnote{https://www.nvidia.com/en-us/autonomous-machines/embedded-systems/jetson-agx-xavier/, https://pixhawk.org/} are adopted as the aerial onboard computer and flight controller. During the real-world UAV tests, RAM usage and GPU VRAM usage are 15.29\% and 3\%, respectively. Additionally, the utilization of GPU and CPU is 46\% and 12.43\% on average. The center location error (CLE) is adopted to evaluate the tracking performance (20 is the success threshold).

\begin{figure}[t]
		\centering	
		\includegraphics[width=1\linewidth]{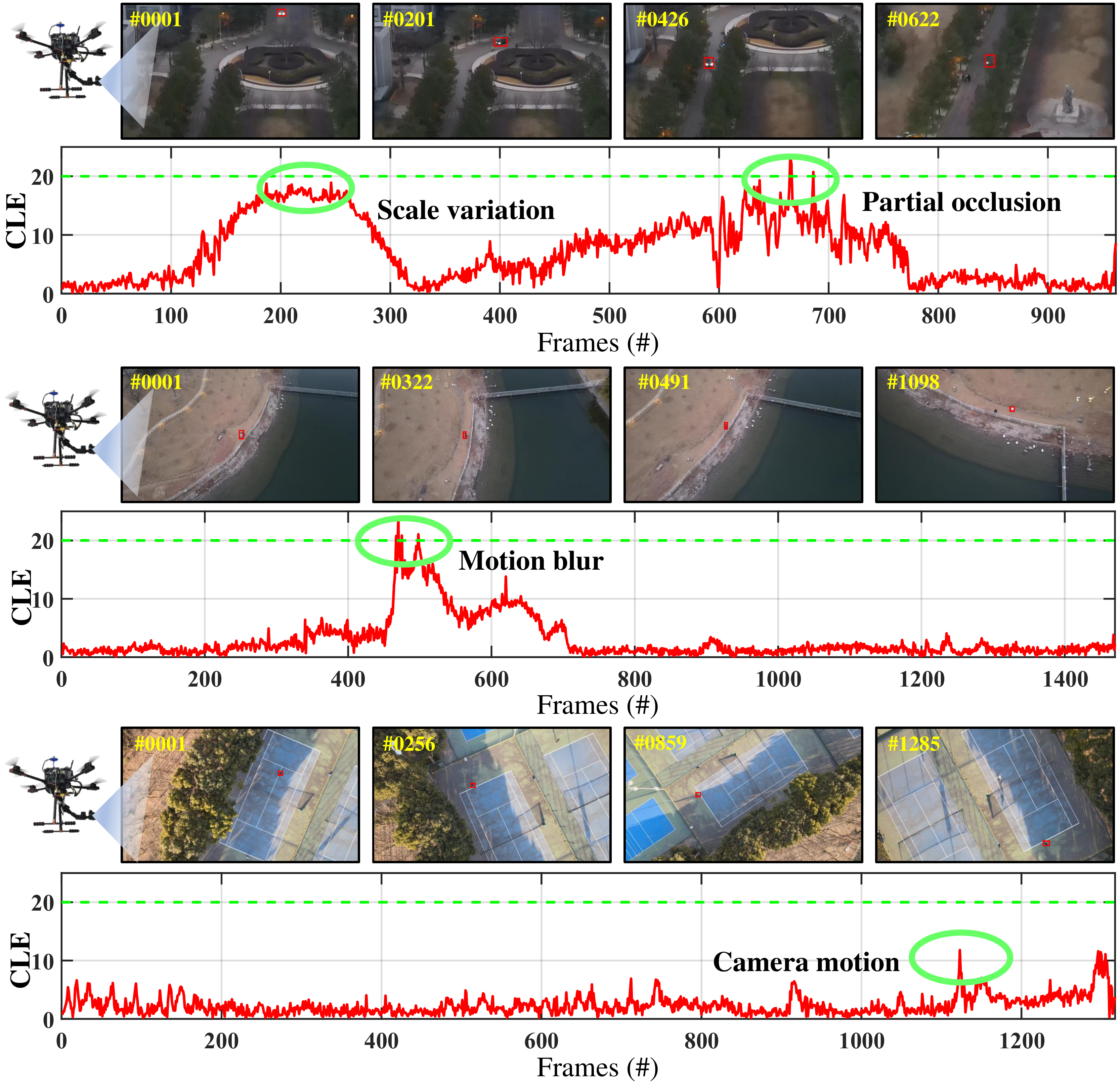}
		\vspace{-20pt}
		\caption{Recording of real-world tests on the embedded platform. The tracking targets are marked with \NoOne{red} while the CLE represents the center location error. To avoid unpredictable disclosure of personally identifiable information, images are processed merely.} 
		\vspace{-20pt}
		\label{fig:v4r}
	\end{figure}

	The special challenges in the real-world tests involve different illumination, scale variation, occlusion, motion blur, and low-resolution scenes. The visualization of our tracking recording of practical UAV is shown in Fig.~\ref{fig:v4r}. When facing partial occlusion and low illumination (the first row), our tracker can maintain impressive stability and robustness via exploiting the consecutive temporal knowledge. Meantime, our tracker also achieves satisfying accuracy when facing motion blur and the occluded object (the second row). Additionally, the visualization of the third row strongly presents the powerful ability of our tracker under camera motion conditions. Finally, our tracker remains at a speed of over 27 FPS during the tests without the acceleration of TensorRT\footnote{https://developer.nvidia.com/tensorrt}. The real-world tests on our practical UAV strongly demonstrate the practicability and feasible deployment ability of our framework. Furthermore, our tracker presents stable and promising tracking performance in complex aerial tracking conditions.

	\section{Conclusion and Discussion}\label{sec:conclusion}
	In this work, we propose a comprehensive framework for introducing temporal contexts into aerial tracking which consists of two perspectives, \textit{e.g.}, feature extraction and similarity refinement. Specifically, in this work, AT-Trans and online TAdaCNN are the first attempts for exhaustively exploring temporal contexts. Besides, attributing to our online updating strategy, unnecessary operations and memory loading are avoided. 
	Extensive experiments on four benchmarks and real-world tests on our UAV demonstrate the effectiveness and efficiency of our framework. 
	We hope that our framework can inspire further research in aerial and even general tracking with temporal contexts.
	
	\noindent\textbf{Potential limitations.} Hindered by the short-term training method, the potential of our framework in very long-term temporal modeling and long-time occlusion is not fully explored. Moreover, the TensorRT and ONNX versions will be developed in our future works.
	
	
	\noindent\textbf{Negative impacts.} Although TCTrack aims to explore temporal contexts comprehensively for aerial tracking, impressive efficiency and effectiveness make it easy to be deployed on UAVs for unauthorized surveillance.

	
	\noindent\textbf{Acknowledgment:} This work is supported by the National Natural Science Foundation of China (No. 62173249) and the Natural Science Foundation of Shanghai (No. 20ZR1460100), by the Agency for Science, Technology and Research (A*STAR) under its AME Programmatic Funding Scheme (Project \#A18A2b0046), by NTU NAP, MOE AcRF Tier 1 (2021-T1-001-088), and under the RIE2020 Industry Alignment Fund – Industry Collaboration Projects (IAF-ICP) Funding Initiative, as well as cash and in-kind contribution from the industry partner(s).

	{\small
		\bibliographystyle{ieee_fullname}
		\bibliography{egbib}
	}
	
\end{document}